\documentclass{article}
\usepackage[final,sglblindworkshop,nonatbib]{neurips_2026}
\author{%
  Nickolay Karlov\\
  Moscow Independent Research\\
  Institute of Artificial Intelligence\\
  \texttt{karlov.n@miriai.org}
  \And
  Alexey Kravatskiy\\
  Moscow Independent Research\\
  Institute of Artificial Intelligence\\
  \texttt{kravatskii.a@miriai.org}
}

\workshoptitle{Neural Network Artifacts as a New Data Modality}
\usepackage[T1]{fontenc}
\usepackage[utf8]{inputenc}
\usepackage[english]{babel}
\usepackage{amsmath,amssymb,amsthm}
\usepackage{graphicx,booktabs,tabularx,array,microtype,xcolor}
\usepackage[round]{natbib}
\usepackage{enumitem,needspace}
\usepackage{float,placeins}
\usepackage[hidelinks]{hyperref}
\usepackage{etoolbox}
\newtheorem{proposition}{Proposition}
\newcommand{\MG}{\widehat d_{\mathrm{MG}}}

\DeclareMathOperator{\median}{median}
\AtBeginEnvironment{table}{\setlength{\tabcolsep}{3pt}}
\AtBeginEnvironment{table*}{\setlength{\tabcolsep}{3pt}}

\newcommand{\appref}[1]{Appendix~\ref{#1}}

\newcommand{\papertitle}{What Can a Time Series of Loss, Weight Norm, or Neuron Activation Reveal About a Neural Network?}
\hypersetup{pdftitle={\papertitle}}

\title{\papertitle}

\begin{document}
\maketitle
\begin{abstract}
It is easy to record the loss or weight norm at each step of neural network training. Can one such time series reveal what is changing inside the network?
For a scalar quantity $x$ recorded over time, we form delay vectors from samples taken at fixed time lags. We estimate the local intrinsic dimension of their distribution with the MacKay--Ghahramani (MG) variant of the Levina--Bickel estimator, which pools log ratios of nearest-neighbor distances across the delay vectors. We call this estimate the \emph{MG of $x$}.
After a recurrent network learns periodic motion, MG from the activation time series of one hidden neuron is close to one, as expected for a periodic orbit.
In ResNet-18, freezing the feature extractor lowers the MG of the weight norm and concentrates weight updates in fewer directions.
In a text variational autoencoder, the MG of reconstruction loss on fixed sentences is lower under settings where the decoder relies less on the latent code.
We also use MG to select among candidate encoders trained on unlabeled MNIST digits with SimCLR, SimSiam, BYOL, and VICReg. For each frozen encoder, we fit a logistic regression classifier to predict digit labels from its features. In initial experiments, the MG of the gradient norm late in training correlated positively with classifier accuracy across encoders. When we selected the encoder with the highest MG of the gradient norm in each held-out run, its test accuracy averaged only 0.41 percentage points below the maximum for that set of encoders.
\end{abstract}

\section{Introduction}\label{sec:intro}
We often finish a training run with a loss curve and only a few saved models. While the optimizer has the weights and gradients in memory, we can also record their norms at little cost. Each adds just one number per logged step. Can these histories reveal when the dynamics of a network become simpler, even while training loss continues to fall?%
  \footnote{\urlstyle{tt}Code is available at
  \href{https://github.com/intsystems/neural-network-dynamics-from-a-scalar-log}{\nolinkurl{github.com/intsystems/neural-network-dynamics-from-a-scalar-log}}.}%

Grokking illustrates what such logs can reveal. Test accuracy can rise long after a network fits its training data. \citet{thilak2022slingshot} observed spikes in training loss as the weight norm of the final layer leveled off after rapid growth. At these transitions, the features entering the classifier changed rapidly and validation accuracy often improved. \citet{notsawo2024predicting} used early oscillations in training loss to predict which runs would later generalize.

Understanding what changes inside the network often draws on more detailed records. To study how a layer represented its inputs during grokking, \citet{notsawo2024predicting} recorded one vector of layer activations per input and estimated the intrinsic dimension of the resulting set of points. Saved weights offer another view of training. \citet{jimenez2026scalar} constructed a scalar series from distances between full weight vectors and used it to measure how rapidly runs from nearby initial weights diverged.

For a quantity already in the log, such as loss or weight norm, we can use the time order of the samples to measure complexity. We form delay vectors whose coordinates are samples taken at fixed time lags \citep{takens1981detecting,sauer1991embedology}. If the fraction of delay vectors within a small radius $r$ of a typical vector scales as $r^d$, the exponent $d$ is the local dimension. We use the MacKay--Ghahramani (MG) modification of the Levina--Bickel estimator, which pools log ratios of nearest-neighbor distances across the delay vectors \citep{levina2004maximum,mackay2005comments}. For brevity, we use \emph{MG of $x$} for the resulting dimension estimate when the delay vectors are formed from the time series $x_1,\ldots,x_N$.

A periodic orbit provides a simple test because its dimension is known. \citet{sussillo2009generating} studied periodic motion in recurrent networks through the joint firing rates of many neurons. We ask whether the activation of one neuron is enough to recover the dimension of the orbit. We train a recurrent network to repeat a figure eight. After training, a single phase specifies the full hidden state, so the orbit is one-dimensional. The estimate from a single hidden neuron is close to one (Section~\ref{sec:force}).

During training, independent measurements help identify what a fall in the MG estimate reflects. In an image classifier, freezing layers and pruning half the weights both lower the MG of the weight norm. The updates concentrate in fewer directions after freezing and spread across more after pruning. In a text autoencoder, regularization settings that reduce the information retained about the input sentence also yield lower MG of reconstruction loss. For a practical test, we select a self-supervised digit encoder using only its gradient-norm history. Choosing the encoder with the highest estimate gives nearly the best linear classification accuracy on average (Section~\ref{sec:ssl}).

\section[Measuring simplification from a scalar log]{Measuring simplification from a scalar log}\label{sec:method}
\subsection{Delay reconstruction}
Suppose we have access only to a scalar log $x_1,\ldots,x_N$. At step $t$ we observe $x_t=\phi(s_t)$, where $s_t$ is the system state and $\phi$ is a scalar measurement function. For example, in minibatch stochastic gradient descent (SGD) with fixed learning rate $\eta$, the state is the parameter vector $s_t=\theta_t$. Given a randomly sampled minibatch $B_t$, the next state is $\theta_{t+1}=\theta_t-\eta\nabla_\theta L_{B_t}(\theta_t)$, where $L_{B_t}$ is the mean loss on that batch. A weight-norm log records $x_t=\|\theta_t\|_2$. With momentum, the state also includes the velocity vector. In our recurrent generators with fixed weights and no external input, the hidden state $s_t=\boldsymbol h_t$ evolves as $\boldsymbol h_{t+1}=F(\boldsymbol h_t)$. For neuron $j$ we record its activation $x_t=\tanh(h_{t,j})$, the value it passes to the rest of the network at step $t$.

\citet{takens1981detecting} gave conditions under which delay coordinates from a single scalar time series reconstruct the state space and dynamics of an autonomous system up to a smooth change of coordinates. Related embedding theorems apply to systems with deterministic or stochastic forcing \citep{stark1999delay,stark2003delay}.

Within each sliding window, index its $W$ consecutive observations from 1 to $W$ and define $z_t=(x_t-\bar x)/\sigma_x$, where $\bar x$ and $\sigma_x>0$ are the window mean and standard deviation. For an integer lag $\tau\ge1$, the window yields $n=W-(E-1)\tau$ delay vectors with $E$ coordinates,
\begin{equation}
y_t=(z_t,z_{t+\tau},\ldots,z_{t+(E-1)\tau})\in\mathbb{R}^E,\qquad 1\le t\le n.
\label{eq:delay}
\end{equation}

\subsection{Pooled local-dimension estimation}\label{sec:mg-neighbors}
\begin{figure*}[!htbp]
\centering
\includegraphics[width=\linewidth]{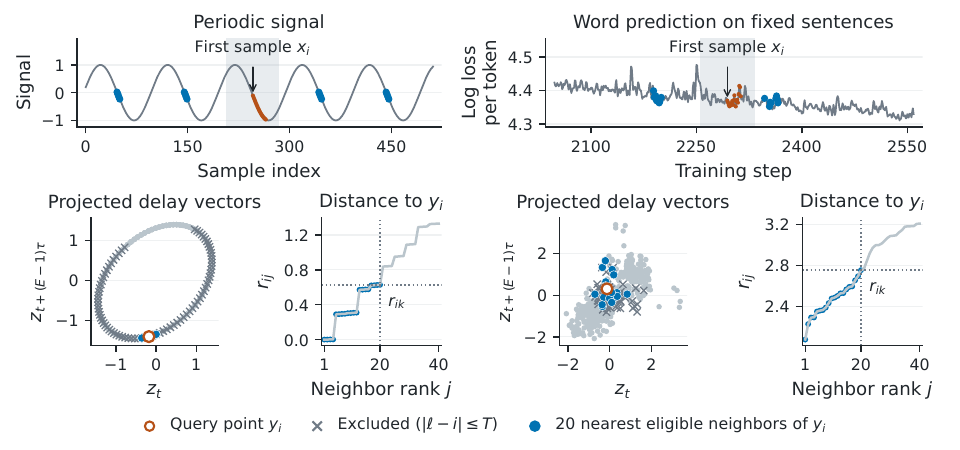}
\caption{Once standardized, the orange samples give the coordinates of one point $y_i$ (open circle below). Blue dots on each series mark neighbor start indices outside the shaded Theiler window.}
\label{fig:method_geometry}
\end{figure*}

We take each delay vector $y_i$ in turn as the query point and search for neighbors within the same sliding window. When $\tau=1$, consecutive delay vectors share $E-1$ samples. Temporal correlations can persist even between disjoint segments. To limit their effect on the neighbor search, we use a Theiler window \citep{theiler1986spurious}. A candidate vector $y_\ell$ is an eligible neighbor of $y_i$ if their starting indices differ by more than $T$ samples, $|\ell-i|>T$. We rank these vectors by Euclidean distance $\|y_\ell-y_i\|_2$ in all $E$ coordinates (Figure~\ref{fig:method_geometry}). We retain the $k$ smallest distances $r_{i1}\le\cdots\le r_{ik}$, where $r_{ij}$ is the distance to the $j$th nearest eligible neighbor of $y_i$.

Suppose the delay vectors sample a smooth $d$-dimensional set with nearly constant density near $y_i$. The expected number of eligible vectors in a small ball of radius $r$ centered at $y_i$ then scales as $r^d$. Doubling the radius encloses about twice as many neighbors on a curve and four times as many on a surface. MG estimates this exponent by comparing the outer radius $r_{ik}$ with each of the $k-1$ inner distances and pooling the log ratios across all $n$ queries,
\begin{equation}
\MG=\frac{n(k-1)-1}
{\displaystyle\sum_{i=1}^{n}\sum_{j=1}^{k-1}
\log(r_{ik}/r_{ij})}.
\label{eq:mg}
\end{equation}
Under the local Poisson model, subtracting one in the numerator makes the estimate unbiased if the sums for different queries are independent (\appref{app:pooled-mg}). A lower estimate corresponds to slower growth of the neighbor count with radius. In figures and tables, the name after @ identifies the scalar time series used to form the delay vectors. For example, MG@loss denotes the MG of loss.

\subsection[Experimental questions]{Experimental questions}\label{sec:mg_conditions}
Consider the long-run fraction of time a trajectory spends within distance $r$ of a typical state. If this fraction scales as $r^d$ at small radii, $d$ is the local dimension of the distribution of visited states. When the delay map preserves distances up to constant factors, MG can estimate this exponent from repeated visits recorded in a scalar log (\appref{app:theory}). The reference dimension is one for a periodic orbit. For a waveform made from several tones, it equals the number of independent phases.

A training trajectory may pass through a region only once. We therefore ask whether a drop in MG coincides with a change we can measure independently, such as weight updates concentrating in fewer directions or a text encoder retaining less information about its input (Table~\ref{tab:experiment-questions}). For loss-spike warnings, we compare MG with log-loss roughness. Roughness is the standard deviation of successive log-loss differences divided by the standard deviation of log loss. We evaluate encoder selection by the resulting classification accuracy and forecaster selection by prediction error (Table~\ref{tab:decisions}).

\begin{table*}[!htbp]
\centering\small
\caption{What MG revealed from a single scalar log in selected experiments.}
\label{tab:experiment-questions}
\setlength{\tabcolsep}{4pt}
\begin{tabularx}{\linewidth}{@{}>{\raggedright\arraybackslash}p{.215\linewidth}>{\raggedright\arraybackslash}p{.19\linewidth}>{\raggedright\arraybackslash}X@{}}
\toprule
Experiment & Scalar log & Result \\
\midrule
\multicolumn{3}{@{}l}{\textbf{Can MG recover the number of independent phases?} (Section~\ref{sec:controlled} and \appref{app:controlled-observables})} \\
Sum of tones & Waveform amplitude & MG was close to one for four harmonics of a single frequency and close to four for four independent frequencies. \\
Diagonal matrix & Frobenius norm or one diagonal entry & As more entries oscillated independently, MG rose for the norm and stayed near one for a single oscillating entry. \\
Linear regression & Training loss & MG rose as more weight directions oscillated independently. \\
Recurrent tone generator & Single-neuron activation & Among generators with four output peaks, MG increased with the number of independent phases. \\
Figure-eight generator & Single-neuron activation & After training, MG was close to one and the full hidden state returned to nearly the same point each period. \\
\addlinespace
\multicolumn{3}{@{}l}{\textbf{How does MG compare with measurements of the weights or latent code?}} \\
Digit classifier (\appref{app:controlled-observables}) & Weight norm or gradient norm & On average, MG from the weight norm was closer to the effective dimension measured from weight covariance than MG from the gradient norm. \\
Small CIFAR-10 classifier (Section~\ref{sec:cnn}) & Weight norm & Training only the final layer reduced MG and concentrated updates in fewer directions. Keeping half the weights at zero reduced MG even as updates spread over more directions. \\
Text variational autoencoder (Section~\ref{sec:vae}) & Reconstruction loss on fixed sentences & Lower MG coincided with less mutual information between input sentences and their latent codes. \\
\addlinespace
\multicolumn{3}{@{}l}{\textbf{Can MG detect a training intervention or warn of a loss spike?} (Sections~\ref{sec:cnn} and \ref{sec:spike-warning})} \\
ResNet-18 & Weight norm & The rule fitted on the small classifier detected every run with a frozen feature extractor. It missed pruning and learning-rate reductions within the evaluation interval. \\
Transformer loss spikes & Logarithm of training loss & The MG rule detected 22 of 23 spikes in advance. The roughness rule detected all 23. Each produced 2.44 false alarms per 10,000 evaluated updates. \\
\addlinespace
\multicolumn{3}{@{}l}{\textbf{How well does MG select encoders or forecasters?} (Section~\ref{sec:ssl} and \appref{app:forecast_utility})} \\
Self-supervised MNIST encoders & Gradient norm & Linear classification accuracy for the encoder with the highest MG averaged 0.41 percentage points below the best in each repeat. \\
Activation forecasts & Single-neuron activation & On held-out networks, choosing by MG or by validation gave similar mean forecast errors. \\
\bottomrule
\end{tabularx}
\end{table*}

\section[Dimension recovery and periodic motion]{Dimension recovery and periodic motion}\label{sec:controlled}

\subsection[Independent oscillations and harmonics]{Independent oscillations and harmonics}\label{sec:generator}
We record the combined amplitude of several sinusoidal tones and form delay vectors from samples at fixed time lags. MG estimates the local dimension of their distribution. The frequencies give us a way to check the estimate. Tones at $f$ and $2f$ repeat together, so knowing the phase of the first fixes the position of both in their cycles. At $f$ and $\sqrt{2}f$, the second tone reaches a different phase on every cycle of the first. There are still two spectral peaks, but describing the motion now takes two independent phases. In our four-tone test, both waveforms have four peaks. MG is close to one for harmonics of a single fundamental and close to four for independent frequencies, as their phase structure predicts (Table~\ref{tab:tone-sums}).

\begin{figure*}[!htbp]
\centering
\includegraphics[width=.98\textwidth]{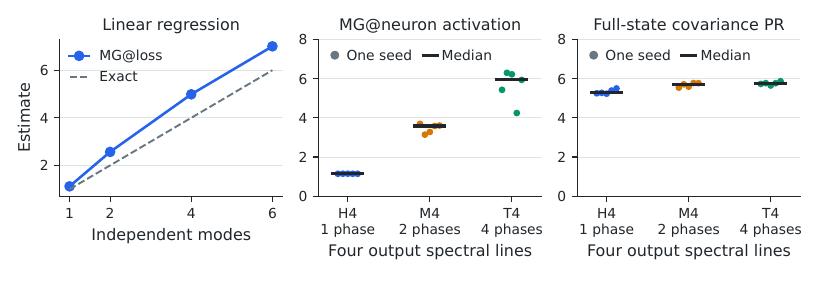}
\caption{Left, MG@loss denotes the estimate from the regression training-loss curve. The dashed line marks agreement with the number of oscillating weight directions. The center shows the MG estimate from the activation time series of one neuron and the right shows PR from all hidden activations for the harmonic (H4), mixed (M4), and independent (T4) four-tone targets. Dots show individual runs and bars show medians.}
\label{fig:components}
\end{figure*}

We can also prescribe how many oscillations contribute to a training loss. In linear regression, we choose the data and initial weights so that, under undamped momentum, the weights oscillate about the least-squares solution at a different frequency along each chosen direction. At every update we record half the mean squared prediction error on the training set. This loss is a weighted sum of the squared displacements from the optimum, so each oscillating direction contributes to the training loss. The MG of training loss rises as more directions oscillate (Figure~\ref{fig:components}, left) and falls when we damp some of them (Table~\ref{tab:regression_variation}).

In a recurrent generator we prescribe the tones at the outputs. A single hidden neuron may respond to only some of the output frequencies. Does the estimate from its activation time series rise when more independent phases are needed to produce the outputs? We fit the output weights to one tone per output using first-order reduced and controlled error (FORCE) learning \citep{sussillo2009generating}. We then fix the weights and let the network run on its own. We verify the tones from the output spectra and form the delay vectors from the activation time series of the first hidden neuron.

We compare networks that start from the same weights and have four spectral peaks in the sum of their outputs. For this neuron, MG is lowest for four harmonics of one fundamental (H4), intermediate for two independent fundamentals and their second harmonics (M4), and highest for four independent frequencies (T4). This ordering holds in every matched repeat (Figure~\ref{fig:components}, center). A generator with three independent tones can give nearly the same single-neuron estimate as one with four (Table~\ref{tab:generator_phases}).

The participation ratio (PR) of the covariance of all hidden activations gives the effective number of principal components. A point moving uniformly around a circle has a PR of two because both axes carry equal variance, even though one phase specifies its position. The effective number of linear directions in the hidden-layer activations is similar for four harmonics of one fundamental and for four independent tones (Figure~\ref{fig:components}, right). \appref{app:controlled-observables} compares MG from a matrix norm and one diagonal entry, and from weight and gradient norms in a digit classifier.

\FloatBarrier
\subsection[Testing for a periodic orbit after training]{Testing for a periodic orbit after training}\label{sec:force}
We train the two outputs of an initially chaotic recurrent network to reproduce the coordinates of a repeating figure eight. A single phase specifies both target coordinates. Because the outputs feed back into the hidden neurons, fitting the output weights with FORCE can make the hidden state periodic. If the hidden state repeats once per target period, the phase also determines the activation of every hidden neuron. Delay vectors from an oscillating neuron should then lie along a curve.

\begin{figure*}[!htbp]
\centering
\includegraphics[width=\textwidth]{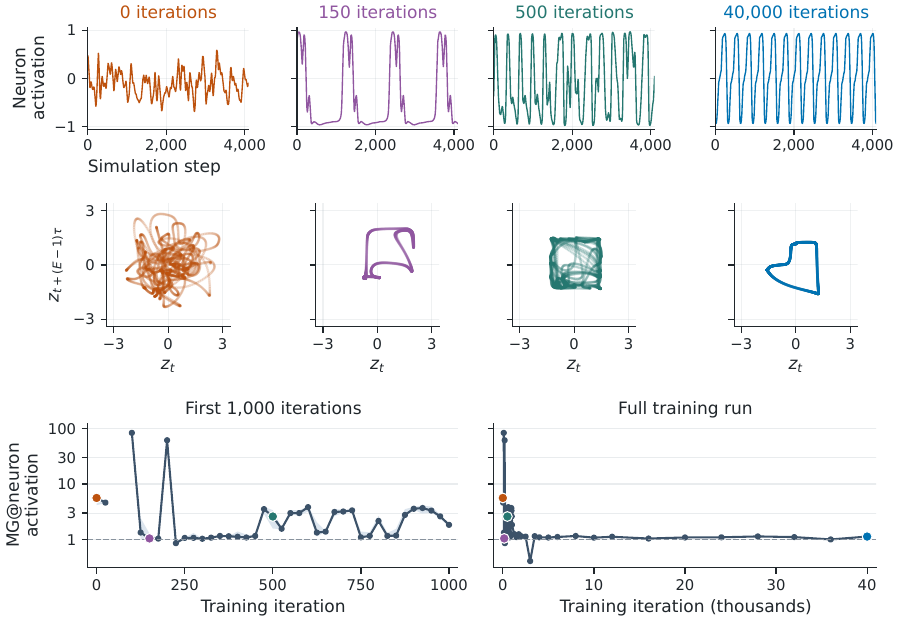}
\caption{Top, activation of the first hidden neuron over time (one simulation step is $0.1$ model time units). Middle, the standardized activation against a delayed copy. Bottom, dots and shading show the median and range of MG across valid windows at each checkpoint. The dashed line marks the dimension of a periodic orbit. At 50 and 75 updates the activation settles to a constant, leaving no valid MG estimate.}
\label{fig:force-log-geometry}
\end{figure*}

At each checkpoint we fix the weights and sample the activation of the first hidden neuron as the network runs without the target. Before training, its activation plotted against a delayed copy forms a broad cloud. At 500 iterations the first few peaks are about one target period apart, but later peaks vary in shape and timing. After training the activation repeats each target period, and pairs of values separated by a fixed lag lie along a closed curve. MG falls close to one (Figure~\ref{fig:force-log-geometry}).

Other hidden neurons can vary across cycles even when the activation of the first neuron repeats. We therefore compare the full hidden state at times about one target period apart. The states nearly coincide in every trained network. To check whether the cycle is stable, we estimate Lyapunov exponents, which measure how quickly small differences between nearby states grow or decay. On a stable cycle a shift in timing persists, while a displacement away from the orbit shrinks. Most runs meet this criterion. In one run with single-neuron MG near one, a second direction contracts too slowly to establish stability by this test (Table~\ref{tab:force-outcome}).

\section{Monitoring and selection in learned systems}\label{sec:applied}

\subsection{Detecting restricted parameter updates}\label{sec:cnn}
We test whether MG can detect when we restrict parameter updates during training. In this sequential change-point detection task, we look for a drop in MG using the weight norms recorded so far. In separate runs of a small convolutional classifier on CIFAR-10 \citep{krizhevsky2009learning}, we freeze layers, prune weights, or reduce the learning rate at a known step. The freezing condition draws on FreezeOut, which stops updating successive layers to reduce training time \citep{brock2017freezeout}. The controls include runs with unchanged training, runs with larger batches, and rescaled or smoothed copies of weight-norm logs from unchanged runs.

The update participation ratio (update PR) measures the effective number of directions in the covariance of the weight changes $\Delta\theta_t=\theta_t-\theta_{t-1}$ within a window. It equals one for variation along a line and $r$ when $r$ orthogonal directions share the variance equally.

\begin{figure*}[!htb]
\centering
\includegraphics[width=.98\textwidth]{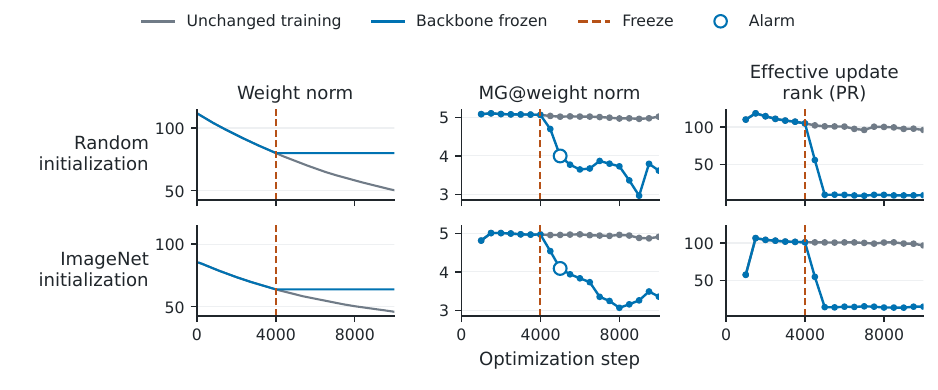}
\caption{Freezing the feature extractor in ResNet-18. Each row compares runs from the same starting weights. MG@weight norm (center) is computed from the weight-norm log at left. At right, a lower update participation ratio (PR) means that fewer directions account for most of the variation among updates. Points are placed at window ends.}
\label{fig:resnet-freeze-trace}
\end{figure*}

In ResNet-18 \citep{he2016deep}, freezing the feature extractor leaves only the final classification layer trainable. Freezing lowers both the MG of the weight norm and the PR of the compressed update vectors (Figure~\ref{fig:resnet-freeze-trace}). Pruning half the weights in the small classifier also lowers the estimate from its weight-norm log. In those runs update PR rises, indicating that the updates spread over more directions (Table~\ref{tab:cnn-graded}).

We fit the rule on one set of classifier runs by varying how many windows enter the earlier and later medians. For each choice we set the threshold above the largest MG drop in the controls. We keep the rule that detects the most interventions. The rule signals when median MG in the latest two windows falls more than $8.83\%$ below the median in the preceding three (Equation~\ref{eq:detector}). On new runs the first alarm must fall within 5,000 updates after the intervention.

For prompt detection in the small classifier, the results favor tracking the mean absolute change between successive weight norms. With interventions at a fixed step, this rule and MG each detected all but one strong intervention without false alarms. The median delay was half as long for absolute changes (Table~\ref{tab:detectors}). When we varied the intervention time, absolute changes detected every event without false alarms, while MG missed events and raised alarms in unchanged runs (Table~\ref{tab:detector-transfer}).

In both architectures, the MG of the weight norm fell after we froze the feature extractor. The threshold fitted on the small classifier detected every freeze in ResNet-18 with either random or ImageNet initialization. It missed pruning and learning-rate reductions there within the scoring interval (\appref{app:applied}). As the pruning example illustrates, interpreting a drop in MG as evidence that updates have become concentrated in fewer directions requires an independent measurement such as update PR.

\subsection[Reconstruction loss and sentence information in a text VAE]{Reconstruction loss and sentence information in a text VAE}\label{sec:vae}
In a text variational autoencoder (VAE), the encoder maps each sentence $x$ to an approximate posterior $q(z\mid x)$ over a sentence-level latent variable $Z$. The decoder predicts each word from a sampled vector $z$ and the preceding words \citep{bowman2016generating}. If the preceding words suffice for prediction, reconstruction loss can stay low even as the decoder becomes less sensitive to $z$. We test whether the MG of reconstruction loss falls as $Z$ carries less information about the sentence.

Starting from the same Penn Treebank model within each repeat \citep{marcus1993building}, we vary the penalty on the Kullback--Leibler (KL) divergence between $q(z\mid x)$ and the standard normal prior. A stronger penalty encourages the posteriors for different sentences to approach the same prior. We also test \emph{free~bits}, which lets the batch-average KL of each latent dimension rise to a threshold before the penalty increases \citep{kingma2016improved}.

We estimate how much information $Z$ carries about the input sentence $X$ using mutual information $I(X;Z)$. For each draw $z$ from $q(z\mid x)$, we compare its density under the posterior for the given sentence with its density under the mixture of posteriors across evaluation sentences. We average the log density ratio over draws and sentences (Equation~\eqref{eq:vae-mi}). To measure decoder sensitivity to $z$, we keep the preceding words fixed and measure how much the predicted word distribution changes when we draw $z$ from the posterior for a different sentence.

\begin{figure*}[!htb]
\centering
\includegraphics[width=.96\textwidth]{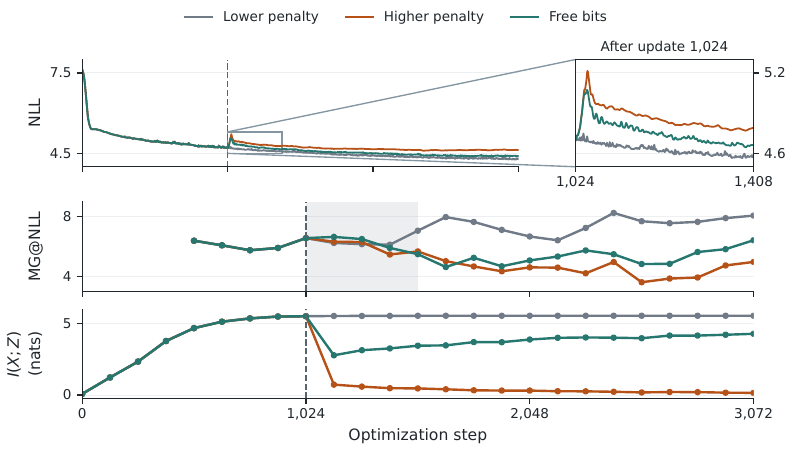}
\caption{Reconstruction loss, the MG estimate from its time series, and sentence information during text VAE training. NLL is the mean negative log-likelihood per token on fixed sentences. $X$ is the input sentence and $Z$ its sampled latent representation. The vertical line marks when the three penalty conditions begin. Shading extends to the end of the first MG window recorded entirely under the new setting.}
\label{fig:vae-loss-log}
\end{figure*}

Reconstruction loss fell overall under every penalty setting in Figure~\ref{fig:vae-loss-log}. Under the stronger KL penalty, giving the decoder a latent vector from another sentence barely changed the predicted word probabilities. With free bits, $Z$ retained more sentence information and the same substitution had a larger effect (Table~\ref{tab:vae-final}).

Across these settings, the less the decoder used $Z$, the lower the MG of reconstruction loss on fixed sentences. Standard deviation and spectral entropy also fell under the stronger penalty (\appref{app:vae-information}). A drop in MG is therefore a reason to check whether the decoder still uses its latent input, even when mean reconstruction loss is improving.

\subsection{Warning of transformer loss spikes}\label{sec:spike-warning}
\begin{figure*}[!htb]
\centering
\includegraphics[width=.92\textwidth]{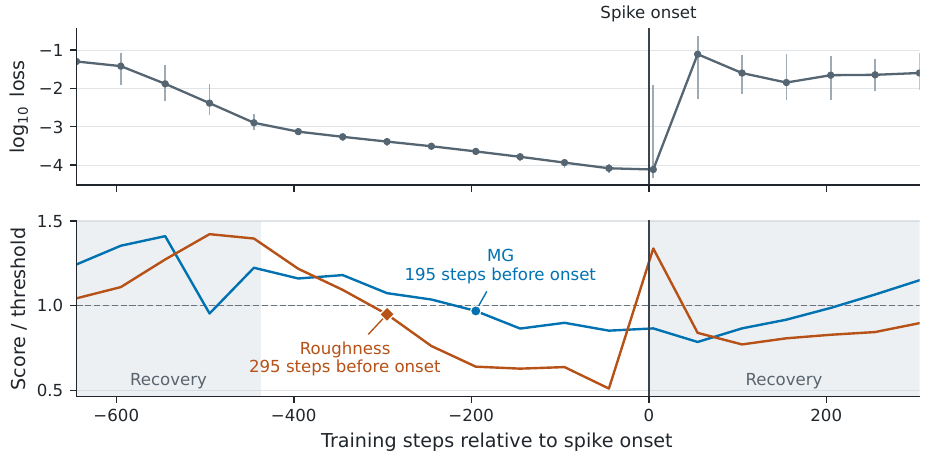}
\caption{Both rules warned before this loss spike in a held-out run. The upper panel shows the mean (points) and range (vertical bars) of log loss between successive checks. Below, each score is divided by its own warning threshold. Outside the shaded recovery periods, a score at or below one (dashed line) triggers a warning. The markers show when each rule first issued a warning in the 500 steps before onset.}
\label{fig:slingshot-warning}
\end{figure*}

When a transformer learns modular addition, can its loss history give advance warning of a spike? We choose warning thresholds on one set of transformer training runs, limiting how often the rules raise false alarms. With those thresholds fixed, a low MG estimate from a recent window of log loss gave advance warning of nearly every spike in held-out runs. A rule based on roughness, which measures step-to-step fluctuations relative to the overall variation in log loss, warned of every spike at the same false-alarm rate (Figure~\ref{fig:slingshot-warning} and Table~\ref{tab:spike_warnings}).

\Needspace{7\baselineskip}
\subsection[Model selection from scalar logs]{Model selection from scalar logs}\label{sec:ssl}
We train encoders on different views of unlabeled MNIST digits, then evaluate each encoder by training a linear classifier on its frozen features. We repeat encoder training with different random seeds. Using one set of repeats, we choose the log and ranking direction that best reproduce the order of classifier accuracies. The resulting rule selects the encoder with the highest MG of its gradient norm late in training. On the remaining repeats, the selected encoders came close, on average, to the best accuracy in each repeat. A measure of feature variation across images ranked the full set of encoders slightly better (Figure~\ref{fig:ssl-probe-scatter} and Table~\ref{tab:encoder_selection}).

Forecasters selected by MG or by validation gave similar mean errors when predicting the activation of a hidden neuron in networks with new initializations (Table~\ref{tab:decisions}). Selecting and fitting a forecaster took a median of 12.2~ms with MG estimated from 128 query points and 45.3~ms with validation (\appref{app:forecast_utility}).

\begin{table*}[!htbp]
\centering\small
\caption{Decisions on held-out runs. A classifier intervention is detected when the first alarm falls in the scoring interval after onset. A loss spike is detected in advance when a warning falls in the interval before onset. Forecast NMSE is normalized mean squared error (mean $\pm$ SD across 20 repeats).}
\label{tab:decisions}
\begin{tabularx}{\linewidth}{@{}>{\raggedright\arraybackslash}Xrrl@{}}
\toprule
Outcome & MG & Baseline & Baseline rule \\
\midrule
Classifier interventions detected & 27/28 & 28/28 & Normalized increments \\
Loss spikes detected in advance & 22/23 & 23/23 & Roughness \\
Accuracy gap (percentage points) & 0.41 & 0.75 & Correlation dimension \\
Forecast NMSE & $0.036\pm0.021$ & $0.041\pm0.032$ & Lowest validation error \\
\bottomrule
\end{tabularx}
\end{table*}

\section{Discussion and conclusion}\label{sec:cost}
A scalar log can reveal when the dynamics of a neural network become simpler. When the recurrent generator learned a repeating cycle, one phase was enough to describe the joint activity of its neurons. Accordingly, the MG of one hidden neuron's activation fell to nearly one.

In the ResNet-18 classifier, we froze the feature extractor and continued training only the final layer. The weight updates then concentrated in fewer directions, and the MG of the weight norm fell. In the text autoencoder, the MG of reconstruction loss was lower under settings where replacing the sentence code had less effect on the predicted words. A scalar training log can thus retain evidence of how a network's weights change during training or how strongly its predictions depend on a particular source of information.

We hope that estimating the local dimension of delay vectors with MG will help reveal changes in the dynamics underlying other scalar time series from machine learning, neuroscience, and robotics.

\section*{AI Use Statement}
The authors formulated the research questions, hypotheses and claims. Generative AI assisted with experimental implementation, debugging, analysis, figures, reports, drafting, translation and editing. The authors verified the reported results, read and revised the manuscript, and take responsibility for the text, claims, code and experiments.

\label{LastMainPage}
\clearpage
\bibliographystyle{plainnat}
\bibliography{references}
\clearpage
\appendix
\onecolumn
\section*{Appendix guide}
\begingroup
\small
\setlength{\tabcolsep}{5pt}
\renewcommand{\arraystretch}{1.18}
\newcommand{\appendixguideentry}[3]{%
  \mbox{\hyperref[#1]{\textbf{\makebox[1.6em][l]{\ref*{#1}}#2}}} & #3\\
}
\begin{tabularx}{\linewidth}{@{}l>{\raggedright\arraybackslash}X@{}}
\toprule[.5pt]
\appendixguideentry{app:theory}{Interpreting MG}{When a dimension estimate from a scalar log reflects the dimension of the motion.}
\appendixguideentry{app:settings}{Reproducibility}{Which signals were recorded and how each experiment was run.}
\addlinespace[.45em]
\appendixguideentry{app:controlled}{Dimension recovery}{Known oscillations in matrices, classifiers, regression and generators.}
\appendixguideentry{app:applied}{Detecting training changes}{How changes in a scalar log compare with weight updates and sentence information in a text VAE.}
\appendixguideentry{app:grokking}{Grokking}{How does the estimate from the weight-norm log change as validation accuracy rises?}
\appendixguideentry{app:walker2d}{Walker2d}{Command smoothing, gait repeatability, and policy selection under actuator lag and noise.}
\appendixguideentry{app:cost}{Computational cost}{Required inputs, computation time, and storage for each diagnostic.}
\addlinespace[.45em]
\appendixguideentry{app:forecast_utility}{Forecasting}{Choosing a forecaster from the activation of a single neuron and estimating MG with fewer neighbor searches.}
\appendixguideentry{app:intervention_utility}{Interventions on MNIST}{Accuracy and computation when stopping training, freezing layers, or pruning neurons.}
\appendixguideentry{app:selection-warnings-transfer}{Selection and warnings}{Image representations, transformer loss spikes and transfer to new tasks.}
\bottomrule[.5pt]
\end{tabularx}
\endgroup
\section[When delay reconstruction supports a dimension estimate]{When delay reconstruction supports a dimension estimate}\label{app:theory}
\subsection{Dimension preservation under delay reconstruction}
When does the dimension measured in a log equal the dimension of the motion that produced it? For a stationary ergodic process, the occupation measure $\mu(A)$ gives the long-run fraction of steps with $s_t\in A$ when the boundary of $A$ has zero measure. Its local dimension at $s$ is
\begin{equation}
d_{\rm act}=\lim_{\epsilon\to0^+}
\frac{\log\mu(B(s,\epsilon))}{\log\epsilon},
\label{eq:active}
\end{equation}
where $B(s,\epsilon)$ is a ball of radius $\epsilon$ in the state-space metric, taken to be Euclidean for the vector states here. We use a single dimension when the limit exists and is constant $\mu$-almost everywhere. For $r$ independent phases on a smooth invariant torus with regular density, the dimension is $r$.

\begin{proposition}[Preservation of local dimension]
Let $\mu$ have local dimension $d$ at $s$. Suppose $F$ is bi-Lipschitz on the support of $\mu$, meaning $c\|x-y\|\le\|F(x)-F(y)\|\le C\|x-y\|$ for all $x,y$ there and constants $0<c\le C<\infty$. The distribution $F_\#\mu$ of reconstructed points then has local dimension $d$ at $F(s)$.
\end{proposition}
\begin{proof}
For sufficiently small $\epsilon$, the inverse image of $B(F(s),\epsilon)$ is bounded on the support between $B(s,\epsilon/C)$ and $B(s,\epsilon/c)$. Their probabilities bound the pushforward probability. The logarithms of $c$ and $C$ contribute vanishing constants after division by $\log\epsilon$, so both bounds have the same limiting exponent $d$.
\end{proof}
Takens gives conditions under which delays of a scalar observation embed the state of a smooth autonomous system \citep{takens1981detecting}. A smooth embedding of a compact smooth manifold supplies the required distance bounds. Sauer et al.\ extend reconstruction to fractal sets \citep{sauer1991embedology}. Stark and colleagues give embedding results for deterministic and stochastic forcing under assumptions on the dynamics and observations \citep{stark1999delay,stark2003delay}. Applying the proposition requires control of distance distortion on the support being measured.

\subsection{Recurrent, transient, and stochastic regimes}
\begin{figure}[H]
\centering
\includegraphics[width=.86\linewidth]{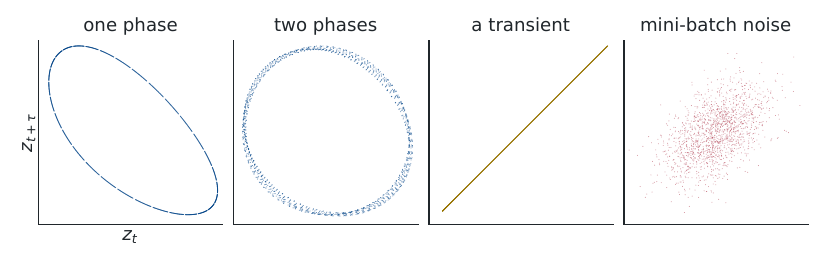}
\caption{Delay reconstructions in the $(z_t, z_{t+\tau})$ plane for periodic, quasiperiodic, transient, and stochastically forced signals.}
\label{fig:shapes}
\end{figure}

In a delay embedding, different states map to different points. To estimate how probability grows with radius from a finite record, we need repeated visits to the same neighborhoods. A periodic orbit supplies such visits. During a transient, successive states may lie along a narrow curve without returning to previously visited neighborhoods. A stochastic input can spread nearby observations apart (Figure~\ref{fig:shapes}). In those cases we interpret MG through independent measurements, as in Section~\ref{sec:applied}.

\subsection{Computing the pooled estimator}\label{app:pooled-mg}
Under the local Poisson model of \citet{levina2004maximum}, the log distance ratios for a query sum to $S_i=\sum_{j=1}^{k-1}\log(r_{ik}/r_{ij})$, which has a gamma distribution with shape $k-1$ and rate $d$. Following \citet{mackay2005comments}, we pool these sums before taking the reciprocal. If the $S_i$ are independent, $S=\sum_i S_i$ is gamma distributed with shape $n(k-1)$ and rate $d$. For $n(k-1)>1$, its reciprocal has mean $\mathbb{E}[1/S]=d/[n(k-1)-1]$, which gives the numerator in Equation~\eqref{eq:mg}. For queries on the same trajectory, neighborhoods can overlap and observations can remain correlated beyond the Theiler window, so the estimate can still be biased.

Repeated patterns in the scalar series can produce coincident delay vectors. A zero neighbor distance makes the log ratio in Equation~\eqref{eq:mg} undefined. We therefore break ties before forming delay vectors and set lower bounds on the distances and on the contribution of each query to the denominator (Table~\ref{tab:mg-numerics}). For the FORCE plot (Figure~\ref{fig:force-log-geometry}), forecast sampling and VAE check scheduling, we omit the estimate for any window in which either bound applies to more than $1\%$ of the corresponding values. Nonfinite or constant records, short windows and too few eligible neighbors yield no estimate. We retain estimates above the embedding dimension $E$.

In a nonconstant window, standardizing $ax_t+b$ gives $\operatorname{sign}(a)z_t$ for any $a\ne0$. At fixed delay and neighbor settings, this preserves all distances in Equation~\eqref{eq:mg} before noise is added. With the Gaussian noise in Table~\ref{tab:mg-numerics}, positive rescalings give the same estimate for the same noise realization, and negative rescalings give the same distribution by symmetry. Rounding can affect nearly constant records.

\section{Configurations and reproducibility}\label{app:settings}
The stride in Table~\ref{tab:mg-settings} is the distance in samples between successive window starts. The first and last samples in one delay vector are $(E-1)\tau$ steps apart. For a query starting at $i$, only vectors starting at $\ell$ with $|\ell-i|>T$ enter the distance ranking (Section~\ref{sec:mg-neighbors}). For the CNN, for example, each vector contains 20 consecutive samples and $T=19$, so none of its eligible neighbors uses any of the same samples. The autocorrelation rule (ACF) uses the longer of $(E-1)\tau$ and the autocorrelation time, up to the cap shown where applicable.

\begin{table*}[htbp]
\centering\small
\caption{Scalar records and MG settings.}
\label{tab:mg-settings}
\begin{tabularx}{\linewidth}{@{}l>{\raggedright\arraybackslash}Xrrrrll@{}}
\toprule
Experiment & Observable & $W$ & $E$ & $\tau$ & $k$ & Exclusion $T$ & Stride \\
\midrule
Four tones & combined amplitude & 8,000 & 20 & 4 & 20 & 76 & 4,000 \\
Digits head & state-based & 8,000 & 20 & 4 & 20 & 76 (cap 150) & 3,000 \\
Regression & training loss / error on one test input & 2,048; 8,192 & 16 & 1 & 10 & 100 & first / last \\
Generator & neuron zero & 8,192 & 20 & adaptive & 20 & ACF; cap 320 & 8,192 \\
FORCE transition figure & neuron zero & 4,096 & 20 & 4 & 20 & 156 & 512 \\
CNN / ResNet-18 & parameter norm & 1,000 & 20 & 1 & 20 & 19 & 500 \\
Text VAE & fixed-set NLL & 512 & 20 & 1 & 20 & 39 & 128 \\
VAE check scheduling & fixed-set NLL & 512 & 20 & 1 & 20 & 39 & 64 \\
Walker2d motion & joint / action scalar & 2,048 & 20 & 8 & 20 & 312 & 1,024 \\
Walker2d selection & action norm & 2,048 & 20 & 8 & 20 & 312 & episode \\
Grokking & parameter norm & 60 & 10 & 1 & 20 & ACF & aligned windows \\
Self-supervised & gradient norm & 1,000 & 20 & 1 & 20 & 19 & final window \\
Spike warning & log loss & 500 & 20 & 1 & 20 & 19 & 50 \\
Forecast selection & single neuron & 2,048 & 20 & adaptive & 20 & ACF; cap 150 & one window \\
MNIST interventions & norm or fixed-set loss & 256 & 10 & 1 & 10 & 19 & candidate updates \\
\bottomrule
\end{tabularx}
\end{table*}

\begin{table}[htbp]
\centering\small
\caption{Numerical safeguards for the pooled MG estimate in Equation~\eqref{eq:mg}.}
\label{tab:mg-numerics}
\begin{tabularx}{\linewidth}{@{}l>{\raggedright\arraybackslash}X@{}}
\toprule
Stage & Rule applied within each window \\
\midrule
Standardized samples & Add independent Gaussian noise with standard deviation $10^{-9}$ to break ties before forming delay vectors. \\
Neighbor distances & Use $\max(r_{ij},10^{-8})$ for each distance before taking log ratios. This keeps the ratios finite for coincident vectors. \\
One query & Sum its $k-1$ log ratios and use at least $10^{-5}$ for that sum before pooling. Equally distant neighbors would give a sum of zero. \\
Whole window & A window is numerically degenerate if more than $1\%$ of its $nk$ neighbor distances reach their floor or more than $1\%$ of its $n$ query sums reach theirs. \\
\bottomrule
\end{tabularx}
\end{table}

\begin{table}[htbp]
\centering\small
\caption{Model architectures by experiment.}
\label{tab:architectures}
\begin{tabularx}{\linewidth}{@{}>{\raggedright\arraybackslash}p{.26\linewidth}X@{}}
\toprule
Task & Model \\
\midrule
Digit classifier & Fixed tanh MLP (64--64), linear classifier in a fixed parameter subspace \\
Synthetic generators and forecasting & FORCE recurrent networks (1,000 neurons for phase recovery, 256 for the periodic-signal and forecast tasks) \\
CIFAR-10 interventions & $3\times3$ ReLU convolutions (16, 32, 32 channels), max pooling after the first two, adaptive average pooling, linear classifier \\
ResNet transfer & ResNet-18, 10 classes. Input convolution: random init.\ $3\times3$, stride 1, no max pooling. ImageNet init.\ $7\times7$, stride 2, max pooling. \\
Text VAE & 48-dimensional shared embeddings, GRU encoder and decoder with hidden size 64, 16-dimensional Gaussian latent variable \\
Grokking & One-layer transformer, width 128, four 32-dimensional heads, MLP width 512, no layer normalization; full-batch comparison uses a width-500 quadratic perceptron \\
MNIST representation selection & 784--256--256 encoder with batch normalization and ReLU; 256--256--64 projector (one variant outputs 16 coordinates); predictor width 64 in the standard SimSiam/BYOL settings \\
MNIST intervention rules & 196--128--128--10 ReLU MLP; pruning narrows both hidden layers \\
Loss-spike warning & Pre-layer-normalized transformer with two layers, width 64, and four heads \\
Walker2d & The actor and critic each have their own pair of 64-unit tanh layers. The actor outputs six commands. \\
\bottomrule
\end{tabularx}
\end{table}

\begin{table}[htbp]
\centering\small
\caption{Classifier intervention and detector settings.}
\label{tab:cnn-protocol}
\begin{tabularx}{\linewidth}{@{}>{\raggedright\arraybackslash}p{.43\linewidth}>{\raggedright\arraybackslash}X@{}}
\toprule
Parameter & Setting \\
\midrule
\multicolumn{2}{@{}l}{\textbf{CIFAR-10 intervention benchmark}} \\
Training images / SGD updates & 10,000 / 10,000 \\
SGD rate / momentum / weight decay & $0.02$ / $0.9$ / $5\cdot10^{-4}$ \\
Batch size / sampling & 64 / with replacement \\
Intervention update & 4,000 \\
Learning-rate divisors & 3, 10, 100 \\
Freezing choices & First two convolutions, all layers before the classifier, or all parameters except the classifier bias \\
Pruned weight fractions & 50\%, 80\%, 95\% \\
Comparison training runs & Unchanged training or batch size $\times4$ \\
Comparison logs from unchanged training & Later samples $\times10$ or a mean over the latest 16 samples \\
Repeats by purpose & 4 to choose detector settings, 4 to test them, 10 to test new intervention times \\
\addlinespace
\multicolumn{2}{@{}l}{\textbf{Detector settings and evaluation}} \\
MG detector & $M=2$, $B=3$, $D_j<-0.08829884$ \\
Number of recent windows ($M$) & 2, 3 or 4 \\
Number of preceding windows ($B$) & 3, 4 or 6 \\
Margin above the largest control drop & 2 percentage points \\
Window ends / post-intervention horizon & $>1{,}500$ / 5,000 updates \\
Strong interventions / comparisons per repeat & 7 / 4 \\
Excluded from the strong set & Learning rate divided by 3, first two convolutions frozen \\
CUSUM baseline window ends & Updates 2,000--3,500 \\
CUSUM scale floor / drift grid / margin & $\max(0.01|\mu_0|,10^{-10})$ / $\{0.25,0.5,1\}$ / $0.5$ \\
\addlinespace
\multicolumn{2}{@{}l}{\textbf{Comparison features}} \\
IAAFT records per window / iterations / trim limit & 2 / 100 / 15\% \\
Permutation entropy & Order 5, delay 1, stable tie sorting, divide by $\log(5!)$ \\
Sample entropy & Length 2, delay 1, Chebyshev distance, tolerance $0.2$ window SDs, no self-matches \\
\addlinespace
\multicolumn{2}{@{}l}{\textbf{Channel inactivity test}} \\
Learning-rate multipliers by duration & 1 update: $\times20$, $\times30$, $\times50$, $\times100$ \\
 & 2 updates: $\times20$ \qquad 3 updates: $\times15$ \\
Images per check / check interval & 500 / 250 updates \\
Earlier / later measurement windows & $[2{,}000,4{,}000)$ / updates 5,000--9,000 \\
Channel loss / network-wide failure & Median inactive fraction rises $\ge0.05$ / training accuracy $<20\%$ \\
Training-loss detector & Median minibatch loss per window. $M=2$, $B=3$, $D_j>-0.0590403$ \\
\addlinespace
\multicolumn{2}{@{}l}{\textbf{Transfer to ResNet-18}} \\
Data / batch size & Full CIFAR-10 without augmentation / 128 \\
Inputs for random / ImageNet initialization & $32\times32$ / $64\times64$ with ImageNet normalization \\
SGD rate / momentum / weight decay & $0.02$ / $0.9$ / $5\cdot10^{-4}$ \\
Repeats per initialization / compressed update size & 3 / 1,024 coordinates for covariance PR \\
\bottomrule
\end{tabularx}
\end{table}

\begin{table}[htbp]
\centering\small
\caption{Text VAE penalty comparisons and check scheduling.}
\label{tab:vae-protocol}
\begin{tabularx}{\linewidth}{@{}>{\raggedright\arraybackslash}X>{\raggedright\arraybackslash}p{.47\linewidth}@{}}
\toprule
Parameter & Setting \\
\midrule
\multicolumn{2}{@{}l}{\textbf{Shared setup}} \\
Training / evaluation sentences & 6,000 / 256 \\
Words before end marker / vocabulary tokens & 8--24 / 2,000 \\
Adam rate / sentences per batch & $10^{-3}$ / 32 \\
Gradient norm limit / dropout & 5 / 0 \\
Decoder context & Actual preceding words \\
\addlinespace
\multicolumn{2}{@{}l}{\textbf{Penalty comparison}} \\
Shared training / continuation ends & 1,024 at $\beta=0.01$ / 3,072 updates \\
KL weights $\beta$ & $0.01$, $1$, $1$ with free bits \\
Free-bits threshold $\lambda$ (nats / coordinate) & 0.5 \\
Reconstruction loss & 16 fixed sentences per update \\
Mutual information and decoder sensitivity & Every 128 updates \\
Initial / final evaluation intervals (updates) & 512--1,024 / 2,048--3,072 \\
MI draws per evaluation sentence & 4 \\
Loss of sentence information (both measures) & $<50\%$ of low-penalty values \\
Retention of sentence information (both measures) & $\ge0.5\times$ early and $\ge2\times$ suppressed values \\
Bootstrap (matched repeats / resamples) & 9 / 20,000 \\
\addlinespace
\multicolumn{2}{@{}l}{\textbf{Penalty schedule and event labels}} \\
Initial training & 1,024 updates at $\beta=0.01$ \\
Schedules and duration & 3 repetitions of 2,048 updates \\
KL weight within each repetition & $0\to1$ over 1,024 updates, then fixed at $1$ \\
Early checks / subsequent check interval & 512--1,024 every 128 updates / every 64 updates \\
Minimum early medians (MI / decoder sensitivity, nats) & $>0.1$ / $>10^{-4}$ \\
Starting values at update 1,024 & Both $\ge0.65\times$ their early medians \\
Loss / recovery ratios to early medians & Both $\le0.50$ / both $\ge0.65$ \\
Event time & Second consecutive qualifying check \\
Minimum time left at onset / maximum detection delay & 512 updates each \\
\addlinespace
\multicolumn{2}{@{}l}{\textbf{Rules for scheduling checks}} \\
Training KL divergence & Mean over the latest 64 updates \\
Scheduled KL weight & $\max\{\beta_t,0.01\}$ at update $t$ \\
Initial checks / extra checks allowed & 1 / 6, 12 (primary), 24 \\
Minimum time between triggered checks & 128 updates \\
Repeats by purpose & 1 to choose the threshold, 9 to test it \\
Threshold grid (primary MG threshold in bold) & 0.025, 0.05, 0.075, 0.1, 0.15, \textbf{0.20}, 0.3, 0.4, 0.5, 0.7, 1, 1.5, 2, 3 \\
Threshold tie-breakers, in order & Fewer unproductive checks, shorter delay, fewer total checks, then larger threshold \\
\bottomrule
\end{tabularx}
\end{table}

\begin{table}[htbp]
\centering\small
\caption{Walker2d training and evaluation settings.}
\label{tab:walker-protocol}
\begin{tabularx}{\linewidth}{@{}>{\raggedright\arraybackslash}X>{\raggedright\arraybackslash}p{.46\linewidth}@{}}
\toprule
Parameter & Setting \\
\midrule
\multicolumn{2}{@{}l}{\textbf{Policy training (PPO)}} \\
Environments / steps per environment per round & 8 / 512 \\
Minibatch size / epochs per round & 256 / 5 \\
Learning rate schedule & $3\cdot10^{-5}\to3\cdot10^{-6}$ \\
Clipping / KL stopping / gradient norm limit & 0.1 / 0.01 / 0.5 \\
Reward discount / advantage estimation factor & 0.99 / 0.95 \\
Entropy / value-loss weights & 0 / 0.5 \\
Normalization statistics & Fixed \\
Steps before / during continued training, summed across environments & 655,360 / 1,048,576 \\
\addlinespace
\multicolumn{2}{@{}l}{\textbf{Command smoothing and state repeatability}} \\
Initial steps discarded / steps recorded at fixed policy & 512 / 4,096 \\
Reference gait (samples / simulator steps per stride) & $152$ / $\approx 152.215$ \\
Reference advance (samples per simulator step) & $(1+\sqrt{2}/1000)^{-1}$ \\
Matched reference-gait penalties and scale & $\lambda=0$ vs.\ $3$ at $\sigma^2=38.4774$ \\
Coordinates for state distance & Omit horizontal position and divide velocities by~5 \\
Lags tested for return error & 20--250 simulator steps \\
Reference gait phase for stride-to-stride variance & Phase 0 \\
Minimum phase crossings per retained episode & 8 \\
Return error and phase-matched variation ratios & Both $\le0.75$ vs.\ no reference reward \\
Minimum baseline return error / phase-matched variation & 0.02 / 0.001 \\
Common completed simulations / reward ratio / extra falls & $\ge 8$ / $\ge90\%$ / $\le1$ \\
\addlinespace
\multicolumn{2}{@{}l}{\textbf{Policy choice under actuator shifts}} \\
Candidate command-smoothing weights & $\lambda\in\{0,0.25,1,4\}$ \\
Training seeds used for evaluation & 4 \\
Additional training steps when policies are saved & 262,144, 524,288, 786,432, 1,048,576 \\
Unperturbed episodes per candidate & 3 \\
Unperturbed steps discarded / recorded & 256 / 2,048 \\
Eligibility (completed episodes / relative reward) & $\ge2$ / $\ge90\%$ of best unperturbed reward \\
Fixed policy choice, if eligible & Final policy at $\lambda=0.25$ \\
Perturbations (mixing weight $\alpha$ / noise SD $\sigma$) & $\{0.05,0.1,0.2\}$ / $\{0,0.02\}$ \\
Perturbed episodes per condition & 5 \\
\bottomrule
\end{tabularx}
\end{table}

\begin{table}[htbp]
\centering\small
\caption{Grokking training and matched windows.}
\label{tab:grokking-protocol}
\begin{tabularx}{\linewidth}{@{}>{\raggedright\arraybackslash}Xll@{}}
\toprule
Transformer setting & Modular addition & Permutation composition \\
\midrule
Task domain & Modulo 113 & $S_5$ \\
Training fraction & 0.30 & 0.50 \\
Weight decay & 1.0 & 0.2 \\
Optimizer updates & 20,000 & 30,000 in 15,000 iterations \\
Runs with / without weight decay & 3 / 1 & 1 / 1 \\
\bottomrule
\end{tabularx}\par\smallskip
\begin{tabularx}{\linewidth}{@{}>{\raggedright\arraybackslash}p{.31\linewidth}>{\raggedright\arraybackslash}X@{}}
\toprule
Shared settings and comparisons & Value \\
\midrule
AdamW rate / $(\beta_1,\beta_2)$ & $10^{-3}$ / $(0.9,0.98)$ \\
Batch size / precision & 256 / float64 \\
Perceptron comparison & Half the modulo-97 pairs, full-batch gradient descent, squared error \\
Window length / center & 60 log samples / time of covariance estimate \\
Training accuracy marking fit & $\ge95\%$ \\
Rise boundaries / minimum windows per interval & First of 5 consecutive logged steps with validation accuracy at or above the cutoff / 3 \\
MG setting grid & $E\in\{4,6,10\}$, $\tau\in\{1,2,4\}$, $k\in\{5,20\}$ \\
Time exclusion before capping & $(E-1)\tau$ steps or the autocorrelation rule \\
\bottomrule
\end{tabularx}
\end{table}

\paragraph{Exact adaptive rules.}
For centered samples $z_t$, define $a(\ell)=\sum_{t=1}^{W-\ell}z_tz_{t+\ell}/\sum_{t=1}^Wz_t^2$. The autocorrelation time $t_a$ is the first lag with $a(\ell)<e^{-1}$, or $W$ if none exists. The autocorrelation routine returns zero when the sample standard deviation is below $10^{-12}$. We compute the adaptive lag from the original window and the autocorrelation exclusion from the standardized, dithered window. The adaptive generator uses
\[
\begin{aligned}
\tau_0&=\max(1,\operatorname{round}(t_a/4)),\\
\tau&=\min\{\tau_0,\max(1,\lfloor W/[8(E-1)]\rfloor)\}.
\end{aligned}
\]
The implementation uses nearest-even rounding. A larger lag $\tau$ leaves fewer complete delay vectors in a window of $W$ samples. The upper bound on $\tau$ limits this loss.

The neighbor search requests $k+2T+1$ candidates to allow for excluded start times and the query itself. To bound this cost, we use $T=\min(T_{\max},T_0)$, where $T_0$ is either a fixed number of steps, $(E-1)\tau$, or $\max\{(E-1)\tau,t_a\}$ for the autocorrelation rule. If $T_{\max}<(E-1)\tau$, an eligible neighbor can share samples with the query.

\paragraph{Window length and recurrence requirements.}
For $T>0$, retaining $k$ eligible neighbors for every query requires $n\ge 2T+k+1$, or $W\ge(E-1)\tau+2T+k+1$. The shared code also checks the input length before reconstruction. It needs at least $E\tau+20$ samples for a fixed lag and $E+20$ for an adaptive lag. For dimension estimation the window must also cover repeated visits within a regime. For detection, longer windows smooth estimates and delay the response.

\paragraph{What is compared within each repeat.}
Within each repeat, CNN and VAE conditions share a starting model, generator tasks share initial weights, and Walker2d rules choose among the same policies. When constructing intervals, we resample all measurements from a repeat together.

\section[Dimension tests and recurrent generators]{Dimension tests and recurrent generators}\label{app:controlled}
We call tones independent when their frequencies $f_j$ are rationally independent. The only integer solution to $\sum_j m_j f_j=0$ is then $m_j=0$ for every $j$ \citep{teschl2012ordinary}.
In each matched pair of four-tone signals (Section~\ref{sec:generator}, Table~\ref{tab:tone-sums}), we use the same amplitudes and initial phase offsets. We choose the fundamental frequency of the harmonic signal so that the two records have the same ratio of the standard deviation of successive differences to that of the signal, $\operatorname{sd}(\Delta x)/\operatorname{sd}(x)$.

\begin{table}[htbp]
\centering\small
\caption{Known dimensions and MG estimates for waveforms made from four tones. We take median MG over the windows in each record (Table~\ref{tab:mg-settings}), then the median over eight paired realizations.}
\label{tab:tone-sums}
\begin{tabular}{@{}lrr@{}}
\toprule
Frequencies in the sum & Independent phases & Median MG \\
\midrule
First four harmonics of one fundamental & 1 & 1.11 \\
Four rationally independent frequencies & 4 & 3.92 \\
\bottomrule
\end{tabular}
\end{table}

\subsection[Oscillations in the diagonal matrix and digit classifier]{Oscillations in the diagonal matrix and digit classifier}\label{app:controlled-observables}
The Frobenius norm combines all the entries of the matrix by summing their squares and taking the square root. Each oscillating diagonal entry varies sinusoidally at one prescribed frequency. As more entries oscillate, the MG of the Frobenius norm rises. The MG of a single oscillating entry stays near one (Table~\ref{tab:matrix-observations}).

\begin{table}[!htbp]
\centering\small
\caption{MG estimates for the Frobenius norm and for one oscillating diagonal entry. For each number of independent oscillations we report the median over four runs. MAE is the mean absolute error against the prescribed dimensions.}
\label{tab:matrix-observations}
\begin{tabular}{@{}lrrrrr@{}}
\toprule
 & \multicolumn{4}{c}{Independent oscillations} & \\
\cmidrule(lr){2-5}
Recorded quantity & 1 & 3 & 5 & 8 & MAE \\
\midrule
Matrix norm & 1.14 & 3.48 & 5.29 & 7.83 & 0.27 \\
One oscillating entry & 1.14 & 1.12 & 1.14 & 1.12 & 3.19 \\
\bottomrule
\end{tabular}
\end{table}

For the classifier, the loss weights set the frequencies, while the optimizer determines how far the weights move in each direction. The MG estimates from the weight-norm log and the loss curve on fixed examples are closer on average to the PR of the weight covariance than the estimates from the gradient-norm log (Table~\ref{tab:synthetic-errors}).

\begin{table}[htbp]
\centering\small
\setlength{\tabcolsep}{2.5pt}
\caption{Digit classifier. For each scalar log, we average the absolute error between the MG estimate and the weight covariance PR over four held-out ranks and four repeats per rank.}
\label{tab:synthetic-errors}
\begin{tabular}{lr}
\toprule
Recorded quantity & Mean absolute error \\
\midrule
Parameter norm & 0.57 \\
Loss on fixed examples & 0.59 \\
Fixed parameter projection & 0.62 \\
Gradient norm & 1.70 \\
\midrule
Pooled over all 10 usable quantities & 0.90 \\
\bottomrule
\end{tabular}
\end{table}

We use the handwritten-digit data distributed with scikit-learn \citep{pedregosa2011scikit}. We train and freeze a multilayer perceptron that extracts features, then train a linear classifier whose weights move within a fixed affine subspace. We randomly assign each training example to a subset before training the classifier. All examples in subset $i$ share the time-dependent coefficient $w_i(t)$ in the cross-entropy loss
\[
w_i(t)=\max\!\left\{0.05,1+0.8\sum_{j=1}^r G_{ji}\sum_{\ell=1}^r A_{j\ell}\sin(2\pi f_\ell t+\phi_\ell)\right\}.
\]
Each column of $G$ sets the coefficients for an image subset. The mixing matrix $A$ balances the effect of the oscillations on early weight updates. We draw phases uniformly on $[0,2\pi)$. For each seed we choose band edges between $1/16$ and $1/8$ cycles per update. A single tone uses their geometric mean. Multiple tones span the band, with interior positions on the log scale set by the fractional parts of square roots of sampled primes.

We compute the gradient norm from the full unweighted training loss in subspace coordinates. At zero learning rate fresh minibatches can still change the loss, while accuracy on fixed examples can take too few distinct values for MG. Table~\ref{tab:learning-oscillator-settings} gives the training settings.

\begin{table}[!htbp]
\centering\small
\caption{Training settings for the digit classifier and regression.}
\label{tab:learning-oscillator-settings}
\begin{tabularx}{\linewidth}{@{}>{\raggedright\arraybackslash}p{.38\linewidth}>{\raggedright\arraybackslash}X@{}}
\toprule
Parameter & Setting \\
\midrule
\multicolumn{2}{@{}l}{\textbf{Digit classifier}} \\
Fixed image subsets / preconditioned step size & 12 / 0.15 \\
Ranks used for calibration / repeats & $\{2,4,6\}$ / 3 \\
Evaluation ranks / repeats & $\{1,3,5,8\}$ / 4 \\
\addlinespace
\multicolumn{2}{@{}l}{\textbf{Dense regression}} \\
Examples / weights & 256 / 32 \\
Initial learning rate / momentum & 1 / 1 \\
Damping update / changed momentum / affected modes & 8,192 / 0.98 / 2 of 4 \\
Start of final window & 12,288 \\
\bottomrule
\end{tabularx}
\end{table}

\subsection[Oscillations in linear regression]{Oscillations in linear regression}
The rows of $X$ are the $N$ training inputs. We choose a full-rank $X$ and fix noiseless targets $y=Xw_\star$. At update $t$ the predictions are $Xw_t$. The scalar logged as training loss is
\[
L_t=\frac{1}{2N}\lVert Xw_t-y\rVert_2^2
   =\frac{1}{2}\sum_{j=1}^{r}\lambda_j q_{j,t}^2.
\]
The second expression writes the loss in the eigenbasis of its Hessian $H=X^\top X/N$. We initialize the weight error $w_0-w_\star$ along $r$ orthogonal directions with distinct eigenvalues $\lambda_j$. Here $q_{j,t}$ is the component of $w_t-w_\star$ along direction $j$. All modes contribute equally to the initial loss. With unit learning rate and unit momentum coefficient, each component follows
\[
q_{j,t+1}=2\cos(\omega_j)q_{j,t}-q_{j,t-1},
\qquad \lambda_j=2(1-\cos\omega_j).
\]
We choose the angular frequencies $\omega_j$ to give $r$ independent phases and rotate the eigenbasis so all weight coordinates vary. Each squared sinusoid has a constant term and an oscillation at twice its original frequency, so $L_t$ depends on all $r$ phases. For comparison we also record $x_{\rm test}^\top(w_t-w_\star)$, the signed prediction error on one fixed test input.

\begin{table}[h]
\centering\small
\caption{Median MG estimates from the training loss and fixed-input prediction error recorded during regression with $r$ known modes.}
\begin{tabular}{rrrr}
\toprule
Modes $r$ & \multicolumn{2}{c}{Training loss} & Fixed-input error \\
\cmidrule(lr){2-3}
 & 2,048 samples & 8,192 samples & 8,192 samples \\
\midrule
1 & 0.90 & 1.12 & 1.01 \\
2 & 2.39 & 2.57 & 2.55 \\
4 & 3.71 & 4.99 & 4.11 \\
6 & 5.88 & 7.00 & 4.98 \\
\bottomrule
\end{tabular}

\end{table}

Estimator settings are fixed across modes. In the damping experiment we reduce momentum along selected eigendirections, so their contributions to the loss decay (Table~\ref{tab:learning-oscillator-settings}). We use the last window of each stationary run and compare the first and last windows in runs with damping. Rotating the eigenbasis changes the fixed-input error. The full-batch loss curve and the MG estimate from that curve stay the same across repeats (Table~\ref{tab:regression_variation}). When the loss becomes nearly constant, standardization can amplify roundoff.

\begin{table}[h]
\centering\small
\caption{MG estimates from the loss and prediction-error records before and after damping two weight oscillations. Each row summarizes the same three regression runs.}
\label{tab:regression_variation}
\begin{tabular}{@{}lrrrr@{}}
\toprule
 & \multicolumn{1}{c}{Training loss} & \multicolumn{3}{c@{}}{Prediction error on one fixed input} \\
\cmidrule(lr){2-2}\cmidrule(l){3-5}
Window & All three runs & Minimum & Median & Maximum \\
\midrule
Before damping & 4.971 & 3.601 & 4.154 & 4.206 \\
After damping & 2.563 & 1.996 & 2.634 & 2.681 \\
\bottomrule
\end{tabular}

\end{table}

\subsection{Learned autonomous generators}
The recurrent generator feeds its output back into the state update,
\[
\begin{aligned}
\boldsymbol h_{t+1}&=0.9\boldsymbol h_t+
0.1(J\tanh\boldsymbol h_t+U\boldsymbol z_t),\\
\boldsymbol z_t&=A^\top\tanh\boldsymbol h_t.
\end{aligned}
\]
Here $J$ couples neurons, $U$ feeds back the output, and $A$ maps activations $\tanh\boldsymbol h_t$ to that output. FORCE fits $A$ to the target while $J$ and $U$ stay fixed. Gain $g$ sets the initial standard deviation of the entries of $J$ to $g/\sqrt n$ for $n$ neurons (Table~\ref{tab:architectures}).

For a task with $q$ outputs, the target for output $j$ is $\sin(2\pi f_j t+\phi_j)$. We draw the initial phase offsets $\phi_j$ once per run and hold them fixed. Table~\ref{tab:generator_phases} gives the frequencies relative to $f_0=(10\pi)^{-1}$. In the task labels, the numeral gives the number of output tones. H denotes one harmonic series, M combines two such series with independent fundamentals, and T uses independent frequencies. For a T target sampled at intervals $\Delta t$, the phases densely fill the $q$-torus if $1,\Delta t f_1,\ldots,\Delta t f_q$ are rationally independent \citep{teschl2012ordinary}.

\begin{table}[htbp]
\centering\small
\caption{Phase structure in learned generators. Entries show mean $\pm$ standard deviation across five runs. We compute the MG estimate from the activation time series of one neuron and PR from the joint activations of all hidden neurons. The MG estimate ranges from $5.00$ to $7.89$ for T3 and from $4.24$ to $6.29$ for T4.}
\label{tab:generator_phases}
\begin{tabular}{llcrr}
\toprule
Task & Target frequencies $f_j/f_0$ & Independent phases & MG & PR \\
\midrule
T1 & $1$ & 1 & $1.14 \pm 0.03$ & $1.66 \pm 0.08$ \\
H2 & $1,\,2$ & 1 & $1.05 \pm 0.18$ & $2.94 \pm 0.09$ \\
T2 & $1,\,\sqrt{2}$ & 2 & $2.80 \pm 0.09$ & $2.98 \pm 0.07$ \\
T3 & $1,\,\sqrt{2},\,\sqrt{3}$ & 3 & $6.37 \pm 1.29$ & $4.36 \pm 0.10$ \\
H4 & $1,\,2,\,3,\,4$ & 1 & $1.151 \pm 0.001$ & $5.33 \pm 0.11$ \\
M4 & $1,\,2,\,\sqrt{2},\,2\sqrt{2}$ & 2 & $3.46 \pm 0.24$ & $5.67 \pm 0.11$ \\
T4 & $1,\,\sqrt{2},\,\sqrt{3},\,\sqrt{5}$ & 4 & $5.62 \pm 0.84$ & $5.75 \pm 0.08$ \\
\bottomrule
\end{tabular}

\end{table}

For each neuron we analyze, we take the median MG over disjoint windows of its activation time series. We compute covariance PR from all activations. For the spectral checks, we count peaks in the sum of the outputs and measure spectral power near the prescribed frequency for each output (Table~\ref{tab:controlled-settings}). Before each Fourier transform we subtract the mean of the $N$ samples and multiply sample $n$ by the symmetric Hann window $w_n=\tfrac12[1-\cos(2\pi n/(N-1))]$, $0\le n<N$. The weights taper to zero at both ends, reducing leakage into distant frequency bins when the record ends partway through a cycle \citep{harris1978windows}.

\begin{table}[!htbp]
\centering\small
\caption{Training and measurement settings for the recurrent generators.}
\label{tab:controlled-settings}
\begin{tabularx}{\linewidth}{@{}>{\raggedright\arraybackslash}p{.38\linewidth}>{\raggedright\arraybackslash}X@{}}
\toprule
Parameter & Setting \\
\midrule
\multicolumn{2}{@{}l}{\textbf{Learned generators}} \\
Base frequency $f_0$ / target amplitude / initial phase & $(10\pi)^{-1}$ / 1 per output / uniform on $[0,2\pi)$ \\
Gain / FORCE steps with recursive least squares / repeats per task & 1.2 / 100,000 / 5 \\
Initial fixed-weight steps discarded / samples kept & 5,000 / 32,768 \\
Simulation step / MG windows per neuron & 0.1 / 4 \\
Sampling for full-state PR & Every fourth activation vector \\
Minimum spectral peak power & 5\% of nonzero-frequency power in summed output \\
Output fidelity & $\ge80\%$ of nonzero-frequency power within 3\% of the prescribed frequency, per output \\
Lyapunov exponents / first steps used / inclusion threshold & 12 largest / 16,384 / $\ge-0.02$ per unit time \\
QR factorization interval / initial propagation discarded (steps) & 10 / 1,000 \\
Noise test (first samples / neurons / tasks) & $\{4{,}096,8{,}192\}$ / 3 / H4, M4, T4 \\
Gaussian noise SD relative to observed SD & $\{0,0.01,0.05\}$ \\
\addlinespace
\multicolumn{2}{@{}l}{\textbf{Transition to periodic motion}} \\
Gain / FORCE steps & 1.5 / 40,000 \\
Largest and next-largest Lyapunov exponents & $|\lambda_1|<0.005$, $\lambda_2<-0.005$ \\
Output and state-return errors & Normalized root mean squared error $<0.1$ for each \\
Initial fixed-weight steps discarded / samples kept & 2,000 / 8,192 \\
\bottomrule
\end{tabularx}\par\smallskip
\begin{tabularx}{\linewidth}{@{}>{\raggedright\arraybackslash}Xrrr@{}}
\toprule
\multicolumn{4}{@{}l}{\textbf{Resolved settings for four-frequency generator logs}} \\
\midrule
Task & $t_a$ & $\tau$ & $T$ \\
\midrule
H4 & 21--29 & 5--7 & 95--133 \\
M4 & 28--40 & 7--10 & 133--190 \\
T4 & 32--40 & 8--10 & 152--190 \\
\bottomrule
\end{tabularx}
\end{table}

Among generators with equally many output frequencies, MG rises with the number of independent phases in every matched repeat. All networks reproduce their prescribed frequencies (Table~\ref{tab:controlled-settings}).

We estimate finite-time Lyapunov exponents by propagating small state perturbations through successive update Jacobians with weights fixed. Periodic $QR$ factorization restores orthonormal perturbations. After the initial transient, summing $\log|R_{ii}|$ and dividing by elapsed time gives the exponent for direction $i$. The number of exponents at or above the cutoff usually matches the number of independent phases, although some harmonic runs have weakly contracting directions close to the cutoff (Table~\ref{tab:controlled-settings}).

Low observation noise preserves the ordering by phase structure. At higher noise the ordering becomes inconsistent.

The prescribed output for the periodic-signal experiment is $(\sin\theta,\tfrac12\sin 2\theta)$, with phase $\theta$ increasing at a constant rate. Before training, nearby hidden states diverge. On a stable periodic orbit, a displacement along the orbit changes only the phase, giving a leading Lyapunov exponent near zero. Perturbations across the orbit decay, giving negative exponents in those directions. The output must also match the prescribed signal after phase alignment, and the hidden state must return close to its starting value after a period (Table~\ref{tab:controlled-settings}). For the latter check we interpolate the shifted states and divide their mean squared return error by total hidden-state variance before taking the square root.

At the 500-iteration checkpoint in Figure~\ref{fig:force-log-geometry}, the first few activation peaks appear regularly spaced, but later peaks differ in shape and timing. In the second half of this fixed-weight simulation, the full hidden state nearly repeats after six target periods, with a normalized return error of $0.04$ at a lag of about $1{,}900$ simulation steps.

\begin{table}[htbp]
\centering\small
\caption{Transition to periodic motion during FORCE training. For each network we take the median of the MG estimates from two nonoverlapping windows of the activation time series from a single neuron. The first two rows give the smallest and largest of these medians across five networks. Figure~\ref{fig:force-log-geometry} uses overlapping windows from one run. In the run that missed the stability criterion, the second Lyapunov exponent was $-0.0022$, above the cutoff of $-0.005$ for contraction across the orbit.}
\label{tab:force-outcome}
\begin{tabular}{lr}
\toprule
Measure & Result \\
\midrule
Single-neuron MG before training & 3.56--5.07 \\
Single-neuron MG after training & 1.08--1.13 \\
Median MG in Figure~\ref{fig:force-log-geometry}, before / after training & 5.62 / 1.13 \\
Runs passing the test for stable periodic motion & 4/5 \\
\bottomrule
\end{tabular}
\end{table}

\section{Applied training procedures and full comparisons}\label{app:applied}
\subsection{CNN and ResNet interventions}
We center the weight-update vectors within each window. If $\lambda_i$ are the eigenvalues of their covariance, the update participation ratio is $(\sum_i\lambda_i)^2/\sum_i\lambda_i^2$.
We compare windows entirely before and after each intervention, then take median changes across repeats. Freezing usually lowers both the MG of the weight norm and the update PR. After moderate pruning they move in opposite directions (Table~\ref{tab:cnn-graded}).

Freezing reinitializes the optimizer in the small CNN. Pruning reapplies its weight mask after each update. The rescaled and smoothed records use logs from unchanged training.

To test whether the power spectrum accounts for the observed changes, we use the iterative amplitude-adjusted Fourier transform (IAAFT) \citep{schreiber1996improved}. We trim segment ends to reduce jumps across the Fourier boundary, then reorder the retained values with IAAFT to approximately preserve their power spectrum. For each run we divide median MG after the intervention by its median beforehand, using the same intervals for controls. Intervention runs usually have lower ratios than controls. Reordering the samples makes this ranking less consistent (Table~\ref{tab:cnn-separation}).

\begin{table*}[t]
\centering\small
\caption{Changes in CNN weight motion across four repeats. The MG estimate uses the weight-norm log.}
\label{tab:cnn-graded}
\begin{tabular}{lrrrr}
\toprule
Condition & \multicolumn{2}{c}{MG@weight norm} & Update PR change & Weights updated \\
\cmidrule(lr){2-3}
 & After & Change & & \\
\midrule
\multicolumn{5}{l}{\textbf{Comparison conditions}} \\
No intervention & 4.32 & +0.4\% & +40.3\% & 100.00\% \\
Batch $\times4$ & 4.85 & +12.8\% & -7.3\% & 100.00\% \\
Rescale the log by 10 & 4.32 & +0.4\% & +40.3\% & 100.00\% \\
Smoothed log & 4.44 & +3.3\% & +40.3\% & 100.00\% \\
\addlinespace
\multicolumn{5}{l}{\textbf{Learning-rate reductions}} \\
Divide by 3 & 4.35 & +0.9\% & -5.4\% & 100.00\% \\
Divide by 10 & 3.95 & -8.3\% & -21.6\% & 100.00\% \\
Divide by 100 & 3.85 & -11.5\% & -20.9\% & 100.00\% \\
\addlinespace
\multicolumn{5}{l}{\textbf{Freezing}} \\
Freeze first two convolutions & 4.27 & -1.8\% & -13.6\% & 65.31\% \\
Train only the final classifier & 3.51 & -18.5\% & -43.5\% & 2.25\% \\
Train only its bias terms & 3.58 & -16.7\% & -68.4\% & 0.07\% \\
\addlinespace
\multicolumn{5}{l}{\textbf{Pruning}} \\
Prune 50\% & 3.80 & -12.7\% & +14.7\% & 50.30\% \\
Prune 80\% & 3.59 & -17.5\% & -10.3\% & 20.48\% \\
Prune 95\% & 3.79 & -11.5\% & -45.3\% & 5.55\% \\
\bottomrule
\end{tabular}

\par\smallskip
\begin{tabular}{@{}lr@{}}
Median MG before intervention (all conditions) & 4.31 \\
Spearman correlation of MG and PR changes & 0.59 \\
\end{tabular}
\end{table*}

\begin{table}[t]
\centering\small
\caption{CNN interventions compared with unchanged training and larger batches. The area under the receiver operating characteristic curve (AUC) measures how often the intervention run in a pair has the lower ratio of later to earlier medians.}
\label{tab:cnn-separation}
\begingroup
\begin{tabular}{lr}
\toprule
Statistic & AUC \\
\midrule
MG & 0.93 \\
MG (IAAFT surrogates) & 0.68 \\
Relative detrended standard deviation & 0.57 \\
Trend crossings & 0.56 \\
Correlation of successive residuals & 0.48 \\
\bottomrule
\end{tabular}
\endgroup
\end{table}

Let $m_j$ be the score for window $j$ ($\MG$ for the MG detector). We compare the median of the latest $M$ scores with the median of the $B$ scores immediately before them. The relative change is
\begin{equation}
D_j=\frac{\median(m_{j-M+1},\ldots,m_j)}
{\median(m_{j-M-B+1},\ldots,m_{j-M})}-1.
\label{eq:detector}
\end{equation}
For example, in one evaluation run we froze all layers before the classifier at update 4,000. At update 5,000 the median MG estimate from the weight-norm log was 4.31 for the previous three windows and 3.76 for the latest two. This 12.8\% drop exceeded the fitted 8.83\% threshold and triggered the first alarm.
The rule raises an alarm when $-sD_j>\delta$, where $s=1$ for a drop and $s=-1$ for a rise. We skip the comparison if any of the $B+M$ scores is nonfinite or the earlier median has magnitude at most $10^{-12}$. MG uses $s=1$. We fit the detector on one set of training runs. For each pair $(M,B)$, we set the threshold two percentage points above the largest drop observed in the unchanged or larger-batch controls. We choose the pair that detects the most interventions in these runs, using shorter delay to resolve ties, then test the fitted rule on held-out runs. An intervention is detected when the first alarm falls within the scoring interval after it. Alarms at or before the intervention are premature. Each comparison record with one or more alarms contributes one false positive (Table~\ref{tab:detectors}).

\subsection[Comparing scalar detection statistics]{Comparing scalar detection statistics}\label{app:baselines}
We compute comparison statistics on the same weight-norm records. Permutation entropy measures the variety of local orderings~\citep{bandt2002permutation}. Sample entropy measures how often similar short sequences recur~\citep{richman2000physiological}. Normalized increments divide the mean absolute change between adjacent samples by the window standard deviation plus $10^{-12}$. Within each window we fit a least-squares line to the weight norm as a function of sample index. Subtracting this line gives residuals $r_1,\ldots,r_W$. Trend crossings count sign changes between consecutive residuals. The lag-1 Pearson correlation compares $(r_1,\ldots,r_{W-1})$ with $(r_2,\ldots,r_W)$. Relative detrended standard deviation divides the residual standard deviation by the magnitude of the original window mean plus $10^{-12}$. Spectral entropy measures how broadly power is distributed across frequencies after detrending. We also test a one-sided cumulative sum (CUSUM) rule~\citep{page1954continuous} that accumulates deviations from an earlier baseline. All statistics use the same run split, event definition, and detector search.

\begin{table}[!htbp]
\centering\small
\caption{Detection of strong interventions in the CIFAR-10 classifier. Delay is the median number of updates among detections.}
\label{tab:detectors}
\scriptsize\setlength{\tabcolsep}{2pt}
\begin{tabularx}{\linewidth}{@{}>{\raggedright\arraybackslash}Xrrr@{}}
\toprule
Statistic & Detections & False positives & Delay \\
\midrule
Normalized increments & 28/28 & 0/16 & 1,500 \\
Sample entropy & 28/28 & 4/16 & 1,500 \\
Permutation entropy & 28/28 & 7/16 & 1,500 \\
Relative detrended standard deviation & 28/28 & 10/16 & 500 \\
Absolute increments & 27/28 & 0/16 & 500 \\
MG & 27/28 & 0/16 & 1,000 \\
Trend crossings & 15/28 & 5/16 & 1,000 \\
Window mean & 4/28 & 0/16 & 1,000 \\
Spectral entropy & 4/28 & 2/16 & 2,750 \\
Correlation of successive residuals & 0/28 & 0/16 & --- \\
\bottomrule
\end{tabularx}

\end{table}

\begin{table}[!htbp]
\centering\small
\caption{False alarms by comparison condition.}
\label{tab:baseline_references}
\scriptsize\setlength{\tabcolsep}{2pt}
\begin{tabular}{lrrrr}
\toprule
Statistic & Unchanged & Batch $\times4$ & Rescaled & Smoothed \\
\midrule
Normalized increments & 0 & 0 & 0 & 0 \\
Sample entropy & 1 & 0 & 1 & 2 \\
Permutation entropy & 3 & 0 & 3 & 1 \\
Relative detrended standard deviation & 2 & 2 & 4 & 2 \\
Absolute increments & 0 & 0 & 0 & 0 \\
MG & 0 & 0 & 0 & 0 \\
Trend crossings & 1 & 2 & 1 & 1 \\
Window mean & 0 & 0 & 0 & 0 \\
Spectral entropy & 0 & 2 & 0 & 0 \\
Correlation of successive residuals & 0 & 0 & 0 & 0 \\
\bottomrule
\end{tabular}

\end{table}

Table~\ref{tab:baseline_rules} gives the fitted directions and thresholds. Both detection rules use the same window statistics.

CUSUM estimates the baseline mean $\mu_0$ and scale $\sigma_0$ from the initial windows (Table~\ref{tab:cnn-protocol}). After this baseline period, it starts at $C_0=0$ and updates $C_j=\max\{0,C_{j-1}+s(\mu_0-m_j)/\sigma_0-\kappa\}$. The sign $s$ sets the direction and $\kappa$ sets the tolerated drift. An alarm occurs when $C_j$ exceeds the threshold $h$. The MG median rule tests for a drop by design. For CUSUM and the other median rules, we choose the alarm direction on the same runs used to set the thresholds (Table~\ref{tab:baseline_rules}).

\begin{table}[!htbp]
\centering\small
\caption{Detector settings fixed before evaluation.}
\label{tab:baseline_rules}
\begin{tabular}{lrrrrrrr}
\toprule
Statistic & \multicolumn{4}{c}{Median comparison} & \multicolumn{3}{c}{Cumulative sum} \\
\cmidrule(lr){2-5}\cmidrule(lr){6-8}
 & $M$ & $B$ & Sign $s$ & Threshold $\delta$ & Sign $s$ & Drift $\kappa$ & Threshold $h$ \\
\midrule
Normalized increments & 4 & 4 & -1 & 0.0289438 & $-1$ & 0.25 & 5.90329 \\
Sample entropy & 2 & 3 & -1 & 0.186699 & $-1$ & 0.25 & 9.86063 \\
Permutation entropy & 2 & 6 & -1 & 0.339038 & $-1$ & 1.0 & 7.79449 \\
Relative detrended standard deviation & 2 & 3 & -1 & 0.305439 & $-1$ & 0.25 & 0.5 \\
Absolute increments & 2 & 3 & 1 & 0.338305 & $+1$ & 1.0 & 19.7929 \\
MG & 2 & 3 & 1 & 0.0882988 & $+1$ & 0.25 & 44.3109 \\
Trend crossings & 2 & 4 & 1 & 0.662857 & $+1$ & 0.5 & 2.31626 \\
Window mean & 2 & 3 & 1 & 0.02 & $+1$ & 0.25 & 0.5 \\
Spectral entropy & 3 & 3 & -1 & 1.29961 & $+1$ & 0.25 & 6.74364 \\
Correlation of successive residuals & 2 & 3 & 1 & 0.0238814 & $+1$ & 0.25 & 0.653369 \\
\bottomrule
\end{tabular}

\end{table}

\begin{table}[!htbp]
\centering\scriptsize\setlength{\tabcolsep}{2pt}
\caption{CNN intervention detection using cumulative sums. Delay is the median among detected interventions.}
\label{tab:cusum}
\begin{tabular}{lrrr}
\toprule
Statistic & Detections & False positives & Delay (updates) \\
\midrule
Relative detrended standard deviation & 28/28 & 4/16 & 500 \\
Normalized increments & 28/28 & 4/16 & 1,000 \\
Permutation entropy & 28/28 & 4/16 & 1,000 \\
Sample entropy & 28/28 & 4/16 & 1,250 \\
Absolute increments & 27/28 & 12/16 & 2,500 \\
MG & 17/28 & 0/16 & 3,500 \\
Trend crossings & 13/28 & 3/16 & 1,000 \\
Spectral entropy & 5/28 & 1/16 & 500 \\
Correlation of successive residuals & 1/28 & 0/16 & 500 \\
Window mean & 0/28 & 0/16 & --- \\
\bottomrule
\end{tabular}

\end{table}

At the fixed intervention time, the MG median rule missed one tenfold learning-rate reduction. With normalized increments, both the median rule and CUSUM detected every strong intervention. Only CUSUM raised alarms on the comparison records (Tables~\ref{tab:detectors} and~\ref{tab:cusum}).

\begin{table}[!htbp]
\centering\small
\caption{CNN detections by time allowed after intervention. Rules needing eight windows can first score at update 5,500, a delay of 1,500 updates.}
\label{tab:horizons}
\begin{tabular}{lrrrrr}
\toprule
Statistic & \multicolumn{4}{c}{Detection within this many updates} & False alarms \\
\cmidrule(lr){2-5}
 & 500 & 1,000 & 2,000 & 5,000 & \\
\midrule
Normalized increments & 0/28 & 0/28 & 21/28 & 28/28 & 0/16 \\
Sample entropy & 8/28 & 13/28 & 27/28 & 28/28 & 4/16 \\
Permutation entropy & 0/28 & 0/28 & 23/28 & 28/28 & 7/16 \\
Relative detrended standard deviation & 25/28 & 27/28 & 27/28 & 28/28 & 10/16 \\
Absolute increments & 15/28 & 16/28 & 26/28 & 27/28 & 0/16 \\
MG & 8/28 & 16/28 & 22/28 & 27/28 & 0/16 \\
Trend crossings & 0/28 & 15/28 & 15/28 & 15/28 & 5/16 \\
Window mean & 0/28 & 4/28 & 4/28 & 4/28 & 0/16 \\
Spectral entropy & 0/28 & 0/28 & 0/28 & 4/28 & 2/16 \\
Correlation of successive residuals & 0/28 & 0/28 & 0/28 & 0/28 & 0/16 \\
\bottomrule
\end{tabular}

\end{table}

A rule needs scores from $B+M$ windows before it can compare the medians. This wait explains some missed early detections in Table~\ref{tab:horizons}. When we shift intervention times, MG sometimes raises an alarm before the intervention. It also raises alarms in runs where training continues unchanged (Table~\ref{tab:detector-transfer}).

\begin{table}[!htbp]
\centering\small
\caption{Detector performance at shifted intervention times.}
\label{tab:detector-transfer}
\begingroup
\begin{tabular}{lrrr}
\toprule
Feature & Detected & Alarms on unchanged runs & Premature alarms \\
\midrule
Absolute increments & 30/30 & 0/10 & 0/30 \\
Normalized increments & 29/30 & 2/10 & 0/30 \\
MG & 19/30 & 5/10 & 6/30 \\
\bottomrule
\end{tabular}
\endgroup
\end{table}

\paragraph{Transfer to ResNet-18.}
We match repeats across random and ImageNet initialization. To estimate update PR, CountSketch \citep{charikar2002finding} assigns each update coordinate a random bin and sign. We sum signed entries in each bin and compute covariance PR from the compressed updates. The transferred detector responds to freezing but misses pruning and learning-rate reductions. Layer norms and random projections produce false alarms.

\begin{table}[!htbp]
\centering\small
\caption{ResNet-18 feature-extractor freezing.}
\label{tab:resnet-freeze}
\begin{tabular}{lrrr}
\toprule
Initialization & Median MG change & Freeze detections & False alarms \\
\midrule
Random & $-27.6\%$ & 3/3 & 0/12 \\
ImageNet & $-30.8\%$ & 3/3 & 0/12 \\
\bottomrule
\end{tabular}
\end{table}

\subsection[Loss of active ReLU channels]{Loss of active ReLU channels}
A ReLU channel is inactive on a set of images if its output is zero for every image. After a brief learning-rate increase, we label channel loss from a rise of at least five percentage points in the median inactive fraction. We then ask whether the MG detector fitted on the earlier CNN interventions flags a drop in the estimate from the weight-norm log in these runs (Table~\ref{tab:cnn-protocol}). We exclude runs with network-wide failure from detection scoring. Runs without channel loss and transformed logs from unchanged training serve as comparisons (Table~\ref{tab:relu-channel-loss}).

\begin{table}[!htbp]
\centering\small
\caption{Detection of newly inactive ReLU channels.}
\label{tab:relu-channel-loss}
\begin{tabularx}{.94\linewidth}{@{}>{\raggedright\arraybackslash}Xrrr@{}}
\toprule
\multicolumn{4}{@{}l}{\textbf{(a) Evaluation records}} \\
\midrule
Record type & Available & Excluded & Scored \\
\midrule
Training runs & 42 & 2 & 40 \\
\quad With channel loss & 23 & 2 & 21 \\
\quad Without channel loss & 19 & 0 & 19 \\
Transformed logs from unchanged training & 12 & 0 & 12 \\
\bottomrule
\end{tabularx}\par\smallskip
\begin{tabularx}{.94\linewidth}{@{}>{\raggedright\arraybackslash}Xrr@{}}
\toprule
\multicolumn{3}{@{}l}{\textbf{(b) Detection results}} \\
\midrule
Statistic & Detections / 21 & False alarms / 31 \\
\midrule
Training-loss comparison & 21 & 8 \\
Relative detrended standard deviation & 21 & 25 \\
Trend crossings & 19 & 7 \\
MG & 12 & 12 \\
Correlation of successive residuals & 0 & 0 \\
\bottomrule
\end{tabularx}
\end{table}
MG can move in opposite directions for channel loss and network-wide failure. Trend crossings detect more channel losses than MG and raise fewer false alarms.

\subsection[Measuring information in the text VAE]{Measuring information in the text VAE}\label{app:vae-information}
Within each repeat, the three training conditions share starting weights, optimizer state, batches, and Gaussian draws. For the batch-average KL divergence $K_j$ of latent dimension $j$, the penalty is $\beta\sum_j K_j$ without free bits and $\beta\sum_j\max(K_j,\lambda)$ with free bits (Table~\ref{tab:vae-protocol}). We log per-token loss on fixed evaluation sentences, reusing the same Gaussian noise when sampling $z$.

For $N$ evaluation sentences and $S$ draws $z_{i\ell}$ from each approximate posterior $q(z\mid x_i)$, we estimate $I(X;Z)$ for a uniformly chosen sentence by
\begin{equation}
\widehat I(X;Z)=\frac{1}{NS}\sum_{i=1}^N\sum_{\ell=1}^S
\log\frac{q(z_{i\ell}\mid x_i)}{N^{-1}\sum_{j=1}^N q(z_{i\ell}\mid x_j)}.
\label{eq:vae-mi}
\end{equation}
Natural logarithms give nats. For the decoder test, we shift encoder distributions by one sentence, keeping preceding words and Gaussian draws fixed. We report decoder sensitivity as the mean of the two KL divergences between the original and reassigned word distributions, averaged over tokens.

\begin{table}[htbp]
\centering\small
\caption{Ratios of median MG estimates from the reconstruction-loss curves in the final VAE evaluation windows, computed within each repeat.}
\label{tab:vae-mg-ratios}
\begin{tabularx}{\linewidth}{@{}>{\raggedright\arraybackslash}Xrr@{}}
\toprule
Conditions (first / second) & Median ratio & 95\% interval \\
\midrule
Higher penalty / lower penalty & $0.517$ & $[0.480,0.542]$ \\
Higher penalty with free bits / higher penalty & $1.286$ & $[1.179,1.418]$ \\
\bottomrule
\end{tabularx}
\end{table}

\begin{table}[!htbp]
\centering\small
\caption{Mutual information, decoder sensitivity and the MG estimate from the reconstruction-loss curve in the final evaluation interval.}
\label{tab:vae-final}
\begin{tabular}{lrrr}
\toprule
Penalty condition & MI (nats) & Decoder sensitivity (nats) & MG \\
\midrule
Lower penalty & 5.545 & 2.087 & 8.129 \\
Higher penalty & 0.327 & 0.021 & 4.172 \\
Higher penalty with free bits & 3.934 & 0.599 & 5.491 \\
\bottomrule
\end{tabular}

\end{table}

Within a run, each final value is the median over the interval in Table~\ref{tab:vae-protocol}, dating MG by its window end. Table~\ref{tab:vae-final} reports medians across repeats. For Table~\ref{tab:vae-mg-ratios} we first divide within repeats, then take the median ratio and resample repeats for its interval. Under the stronger penalty, the loss curve on fixed sentences gives a lower MG estimate in every repeat. Its standard deviation and spectral entropy also fall in every repeat. The decrease in the estimate is less consistent for minibatch loss or longer delays.

\subsection[Scheduling checks for loss and recovery of sentence information]{Scheduling checks for loss and recovery of sentence information}
As the KL penalty cycles, frequent measurements of mutual information and decoder sensitivity show when the model loses or regains sentence information. We use these transitions to evaluate schedules with a limited number of checks.

We check both measures when $|\log(m/m_{\rm last})|$ reaches a threshold. Here $m$ is the MG estimate from the current window of the reconstruction-loss curve and $m_{\rm last}$ is its value at the previous check, initially the median over the early windows. For each budget we choose the threshold that detects the most transitions in one calibration run. In the evaluation runs, we count each transition once if a scheduled check confirms the new state within the allowed delay from onset and before the next transition (Table~\ref{tab:vae-protocol}).

At the smallest budget, MG-triggered checks detect more transitions than periodic checks. At the primary budget, periodic checks detect more even when both schedules use the same number of checks (Table~\ref{tab:cyclic}).

\begin{table}[!htbp]
\centering\small
\caption{VAE information-loss and recovery events detected out of 41.}
\label{tab:cyclic}
\begin{tabular}{lrrr}
\toprule
Method & \multicolumn{3}{c}{Extra checks allowed} \\
\cmidrule(lr){2-4}
 & 6 & 12 & 24 \\
\midrule
Periodic checks & 12 & 38 & 40 \\
MG & 18 & 28 & 36 \\
Spectral entropy & 17 & 28 & 35 \\
Standard deviation & 13 & 20 & 38 \\
Scheduled KL weight & 18 & 18 & 26 \\
Training KL divergence & 8 & 13 & 31 \\
\addlinespace
Periodic with the same number of checks as MG & --- & 34 & --- \\
\bottomrule
\end{tabular}

\end{table}

\FloatBarrier
\section[Grokking]{Grokking}\label{app:grokking}
Grokking is the rise in accuracy on held-out examples long after a model fits its training set. We use the one-layer transformer and addition modulo 113 from \citet{nanda2023progress}, training with minibatches. We also train this architecture to compose permutations in $S_5$, a task from \citet{power2022grokking}. Both transformer tasks include a control with zero weight decay. For the full-batch experiments, we train the quadratic perceptron of \citet{gromov2023grokking} by gradient descent on modular arithmetic tasks, including polynomials from \citet{doshi2024grokking} (Table~\ref{tab:grokking-protocol}).

During training we record the weight norm and compute class scores (logits) on a fixed set of inputs. Before compressing the weight and logit vectors, we center the logits across classes for each input and scale the resulting vector to unit length. Within each window we fit a least-squares line $a_j+b_j t$ to each compressed coordinate $y_{tj}$, where $t$ indexes samples. We compute covariance PR from the residuals $y_{tj}-a_j-b_j t$ and compare it with PR from the original coordinates. For the detrended MG comparison (Table~\ref{tab:grokking-drop}), we likewise subtract the line fitted to the scalar weight norm in each window before constructing its delay vectors.

\begin{figure}[H]
\centering
\includegraphics[width=.96\linewidth]{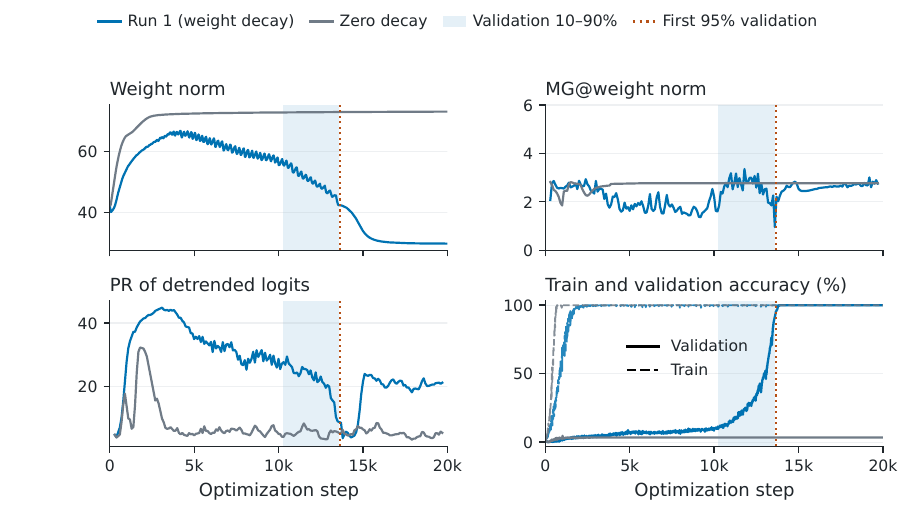}
\caption{Modular addition with and without weight decay. For the run with weight decay, shading marks the rise in validation accuracy from 10\% to 90\%. The dotted line marks the first logged step at which validation accuracy is at least 95\%.}
\label{fig:grokking-matched-traces}
\end{figure}

In the modular-addition example, MG rises as validation accuracy improves and falls, along with logit PR, near the end of the rise. MG also varies in the control while validation accuracy stays low (Figure~\ref{fig:grokking-matched-traces}).

\begin{figure}[H]
\centering
\includegraphics[width=.96\linewidth]{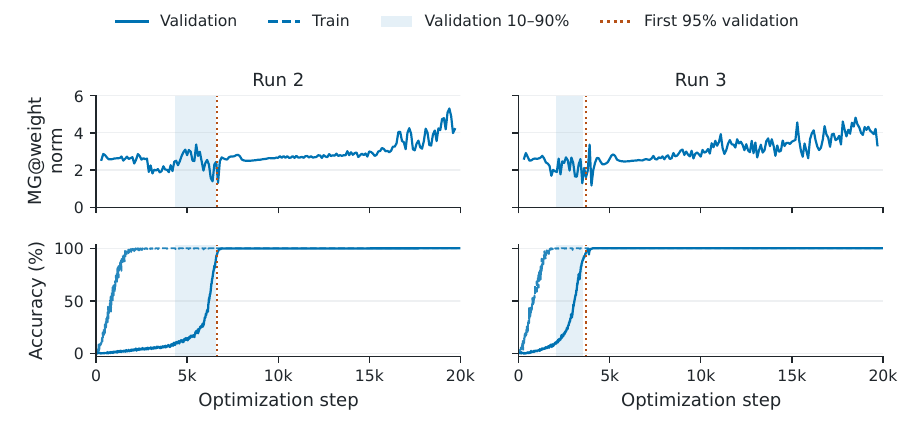}
\caption{MG estimates from the weight-norm log and accuracy in two further modular-addition runs with weight decay.}
\label{fig:grokking-other-runs}
\end{figure}

Of the two further runs with weight decay, one also has higher MG during the rise than on either side. In the other, median MG is similar during and after the rise, and the gap between training fit and the rise is too short for a complete window (Figure~\ref{fig:grokking-other-runs}).

Across transformer runs, we locate the rise from sustained crossings of validation accuracy. We compare the median MG estimate from the weight-norm log during the rise with the medians in equally long adjacent intervals, using only complete windows after training fit (Table~\ref{tab:grokking-protocol}). A peak during the rise is uncommon and depends on how the rise is defined (Table~\ref{tab:grokking-rise}).

\begin{table}[!htbp]
\centering\small
\caption{Runs in which median MG is higher during the rise in validation accuracy than on either side.}
\label{tab:grokking-rise}
\begin{tabular}{@{}lrr@{}}
\toprule
Accuracy interval & Runs with higher MG & Comparable runs \\
\midrule
10--90\% & 2 & 6 \\
20--80\% & 1 & 6 \\
\bottomrule
\end{tabular}
\end{table}

We also test retrospectively for a brief dip in the MG estimate from the weight-norm log. Relative to the generalization step, we divide median MG from steps $-3{,}000$ to $-1{,}000$ by minimum MG from $-1{,}000$ to $+2{,}000$, using matching steps in the controls. A larger ratio indicates a deeper dip. With most estimator settings, even the shallowest dip among generalizing runs is deeper than the deepest control dip. After detrending, none of the settings meets this criterion (Table~\ref{tab:grokking-drop}).

\begin{table}[H]
\centering\small
\caption{Estimator settings for which the MG estimate from the weight-norm log dips more deeply in every grokking run than in any control.}
\label{tab:grokking-drop}
\begin{tabular}{lrr}
\toprule
Weight-norm record & Settings meeting criterion & Usable settings \\
\midrule
Original & 24 & 26 \\
After removing a linear trend & 0 & 26 \\
\bottomrule
\end{tabular}
\end{table}

Covariance PR is also sensitive to drift and correlation time \citep{antognini2018pca}. Roughness changes more than MG. The full-batch perceptron shows no comparable change in MG.

\FloatBarrier
\section[Walker2d]{Walker2d}\label{app:walker2d}
\subsection[Command smoothing and gait repeatability]{Command smoothing and gait repeatability}
A Walker2d policy controls the joints of a simulated biped. Starting from a walking policy trained by proximal policy optimization (PPO) \citep{schulman2017proximal,todorov2012mujoco}, we continue training with a penalty on changes in successive commands.

The smoothing objective is
\[
L=L_{\rm PPO}+\frac{\lambda}{6}\mathbb E\|\bar a_\theta(s_{t+1})-\bar a_\theta(s_t)\|_2^2,
\]
where $\bar a_\theta(s)$ is the mean command vector clipped to $[-1,1]$ per actuator. We average over consecutive steps within episodes and compare the MG estimate from the command-norm time series with the corresponding estimate at $\lambda=0$ in the same repeat.

Does a more regular record of commands imply a more repeatable stride? In the reference-gait experiment, both policies start from the same weights and receive the sine and cosine of a prescribed stride phase. We subtract $\lambda(1-\exp[-e_t^2/(2\sigma^2)])$ from the reward, where $e_t^2$ is squared distance to the interpolated state at that phase. We omit horizontal position and rescale velocities when computing the distance (Table~\ref{tab:walker-protocol}).

Return error compares the state series with a shifted copy. We use the lag giving the smallest mean squared difference and normalize by twice the total state variance. A second measure asks how much the state varies when each stride returns to the same phase. We divide the variance of the interpolated states at that phase by total state variance. Success requires both measures to improve while reward and completion are preserved (Table~\ref{tab:walker-protocol}). Command smoothing consistently lowers MG, while only some policies trained to follow a prescribed stride meet the state criteria (Table~\ref{tab:walker-motion}).

\begin{table}[H]
\centering\small
\caption{Command smoothing and state repeatability in Walker2d.}
\label{tab:walker-motion}
\begin{minipage}[t]{.43\linewidth}
\centering
\begin{tabular}{@{}lrr@{}}
\multicolumn{3}{@{}l}{\textbf{Command smoothing}} \\
\toprule
Penalty $\lambda$ & Median MG ratio & Lower MG \\
\midrule
$0.25$ & 0.801 & 5/5 \\
$1$ & 0.604 & 5/5 \\
\bottomrule
\end{tabular}
\end{minipage}\hfill
\begin{minipage}[t]{.51\linewidth}
\centering
\begin{tabular}{@{}lr@{}}
\multicolumn{2}{@{}l}{\textbf{Training with a reference gait}} \\
\toprule
Outcome & Repeats \\
\midrule
State and performance criteria met & 2/5 \\
Lower MG in every reconstruction & 1/5 \\
\bottomrule
\end{tabular}
\end{minipage}
\end{table}

\subsection{Selecting policies for actuator shifts}\label{app:rl_selection}
Can unperturbed trials identify a policy that retains reward when actuators respond late or noisily? Each rule chooses from the same candidates using trials with direct actuation. We evaluate the chosen policies in new trials with actuator lag and noise (Table~\ref{tab:walker-protocol}).

We test the hypothesis that a policy with more regular commands will retain reward under actuator shifts. On unperturbed trials we compute mean reward over all episodes, including incomplete ones. We keep policies that finish at least two of three episodes and reach 90\% of the best mean reward. For each eligible policy we take the median MG estimate from its command-norm records over completed episodes and select the policy with the lowest median. We fall back to the highest unperturbed reward if no policy qualifies or none of the eligible policies has a valid estimate. The fixed choice from the preliminary run also falls back to reward when ineligible. The random rule samples eligible policies uniformly. Entropy, normalized squared increments and recurrence use the same scalar record. Increments average squared successive differences divided by twice the variance. Recurrence minimizes the same difference over the lags used in the state-return test (Table~\ref{tab:walker-protocol}). Action smoothness averages squared changes across actuators.

In the perturbed trials, we compute $v_t=(1-\alpha)u_t+\alpha v_{t-1}+\epsilon_t$ from the current command $u_t$ and the previous filtered value. We start with $v_{-1}=0$ and add independent Gaussian noise of standard deviation $\sigma$ per component. The simulator receives $v_t$ clipped to $[-1,1]$. The recursion retains the unclipped value, including during the initial steps discarded from scoring. Mean reward is the total divided by the planned number of steps, with zero reward after a fall.

\begin{table}[ht]
\centering\small
\caption{Walker2d reward and completion under actuator delay and noise.}
\label{tab:rl_selection}
\begin{tabular}{lrr}
\toprule
Selection rule & Mean reward & Completed episodes \\
\midrule
Spectral entropy & 4.480 & 97.5\% \\
Scalar recurrence & 4.398 & 95.0\% \\
MG (action norm) & 4.313 & 90.8\% \\
Fixed policy choice & 4.310 & 90.8\% \\
Random eligible policy & 4.286 & 90.9\% \\
Highest unperturbed reward & 4.005 & 80.0\% \\
Normalized squared increments & 3.064 & 49.2\% \\
Action smoothness & 2.999 & 47.5\% \\
\addlinespace
Best using perturbed trials & 4.671 & 100.0\% \\
\bottomrule
\end{tabular}

\end{table}

MG usually improves on selecting by unperturbed reward and gives nearly the same mean reward as the fixed policy. Spectral entropy yields the highest mean reward among rules using unperturbed trials. The final row gives the best eligible choice if perturbed rewards are known (Table~\ref{tab:rl_selection}).

\section{Timing procedures and information requirements}\label{app:cost}
We report the time to record a signal and the time to compute a diagnostic from it. Direct checks require access to the state update and hidden states for Lyapunov exponents, or to the encoder and decoder to measure mutual information and decoder sensitivity in the VAE.

\subsection{Timing environments}\label{app:hardware}
The FORCE and forecast timing runs use an Intel Core i5-12500H CPU with one numerical thread. The VAE scheduling test uses two PyTorch threads and one estimator thread. Walker2d timing uses one numerical thread. Hardware metadata are incomplete for the CNN and the VAE scheduling test.

\subsection{Benchmark-specific timing and storage}

\begin{table}[htbp]
\centering\small
\caption{Analysis time on saved records. FORCE times are medians before and after training, in that order.}
\label{tab:cost}
\begin{tabular}{lrlr}
\toprule
Setting & MG & Comparison calculation & Time \\
\midrule
CNN, 1,000-point window & 9.5\,ms & --- & --- \\
FORCE, 256 neurons & 0.835 / 0.598\,s & Full Lyapunov spectrum & 12.006 / 11.551\,s \\
VAE, 512-point window & 12.89\,ms & MI and decoder sensitivity & 209.32\,ms \\
\bottomrule
\end{tabular}
\end{table}

\begin{table}[htbp]
\centering\small
\caption{Storage and timing by operation.}
\label{tab:record-cost}
\begin{tabularx}{\linewidth}{@{}p{.20\linewidth}Xl@{}}
\toprule
Experiment & Quantity or operation & Measured cost \\
\midrule
CNN & Store the 32-bit scalar log & 39.06\,KiB \\
& Store the sequence of weight vectors & 559.46\,MiB \\
\addlinespace
FORCE & Simulate and log the activation of one neuron before training & 0.102\,s \\
& Simulate and log the activation of one neuron after training & 0.166\,s \\
& Largest Lyapunov exponent before training & 0.198\,s \\
& Largest Lyapunov exponent after training & 0.153\,s \\
\addlinespace
Generators & MG on four windows & 8--23\,s \\
& Principal components of all activations & $\sim0.2$\,s \\
\addlinespace
VAE scheduling & One evaluation of loss on fixed sentences & 7.134\,ms \\
& One joint evaluation of MI and decoder sensitivity & 209.323\,ms \\
& All loss evaluations at the primary budget & 51.14\,s \\
& Total estimate with MG at the primary budget & 54.68\,s \\
\addlinespace
Walker2d selection & Simulate three unperturbed episodes per policy and log actuator commands & 5.52\,s \\
& Trials under actuator delay and noise & 44.7\,s \\
& MG & 0.516\,s \\
& Spectral entropy & 0.00014\,s \\
& Scalar recurrence & 0.0062\,s \\
\bottomrule
\end{tabularx}
\end{table}

Once the data are recorded, MG is faster than the full Lyapunov spectrum on FORCE states but slower than principal components on generator activations (Tables~\ref{tab:cost} and~\ref{tab:record-cost}).

Multiplying each median time by its frequency of use shows that evaluating loss on fixed sentences accounts for most VAE monitoring time. In Walker2d, scalar statistics reuse the recorded commands, while testing gait stability requires further simulations from perturbed states.

\section{Forecast selection and sampled-query estimation}\label{app:forecast_utility}
\subsection[Forecast model choice]{Forecast model choice}\label{sec:forecast_utility}
Each forecast task uses an activation time series from one neuron. The forecaster receives an observed segment, and we reserve its continuation for scoring. The series come from untrained networks and networks trained to produce the oscillations in Table~\ref{tab:generator_phases}.
We fit a regression tree using forecast errors from the candidate models on training tasks with known continuations. On a new task, the tree uses the MG of the observed activation segment to select a forecaster.

\begin{table}[htbp]
\centering\small
\caption{Forecast benchmark design. Network sizes are in Table~\ref{tab:architectures}.}
\label{tab:forecast-design}
\begin{tabularx}{\linewidth}{@{}>{\raggedright\arraybackslash}X>{\raggedright\arraybackslash}p{.53\linewidth}@{}}
\toprule
Parameter & Setting \\
\midrule
\multicolumn{2}{@{}l}{\textbf{Networks and recorded series}} \\
Trained / untrained recurrent gain & 1.2 / 1.5 \\
Simulation step / FORCE training steps & 0.1 / 40,000 \\
Recursive least-squares matrix & Identity initialization, update every 2 steps \\
Recorded activation & $\tanh(h_{t,2})$ (zero-based neuron 2) \\
Initial steps discarded / samples recorded & 5,000 / 8,192 \\
Observed / forecast samples & 4,096--6,143 / 6,144--6,175 \\
\addlinespace
\multicolumn{2}{@{}l}{\textbf{Candidate forecasts}} \\
Fixed forecasts & Last value, observed mean, or repeat at lag 314 \\
Ridge or 5-nearest-neighbor regression (input samples) & 8 or 48 \\
Fourier ridge (random features / input samples) & 64 / 48 \\
Ridge penalty / intercept penalty / Fourier bandwidth & 1 / $10^{-8}$ / $1/48$ \\
Bounds on standardized forecasts & $\pm10$ \\
Repeat last value if observed SD is below & $10^{-10}$ \\
Minimum variance in NMSE denominator & $10^{-12}$ \\
\addlinespace
\multicolumn{2}{@{}l}{\textbf{Series-statistic settings}} \\
Time points for the fitted trend & Equally spaced on $[-1,1]$ \\
Shifts tested for recurrence (samples) & 6--288 \\
\addlinespace
\multicolumn{2}{@{}l}{\textbf{Model selection and timing}} \\
Repeats used to fit the selection rule & 18, grouped by initialization into 3 folds \\
Decision tree (depths / minimum leaf size) & $\{2,3\}$ / 8 \\
Random forest (trees / depths / minimum leaf size) & 100 / $\{3,6\}$ / 5 \\
Validation forecasts per candidate & 4 successive forecasts of 32 samples each \\
Query vectors sampled per window & 128 in the main test, 64 and 256 in sensitivity tests \\
Query vectors processed at a time & Up to 64 \\
Timing CPU / numerical threads & Intel Core i5-12500H / 1 \\
Common inputs / timed repeats per input & 12 / 7 \\
\bottomrule
\end{tabularx}
\end{table}

\begin{figure}[!htbp]
\centering
\includegraphics[width=\linewidth]{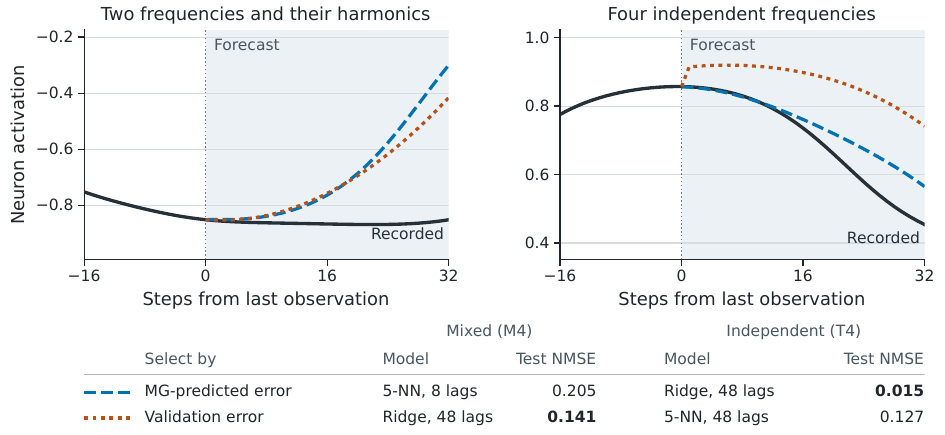}
\caption{Forecasts selected by MG (blue dashes) and validation (orange dots). The solid line shows the measured activation of the hidden neuron. Test NMSE is the normalized mean squared error over the shaded forecast interval. On the left both forecasts rise while the observed values stay nearly constant. Validation selects the model with the smaller rise. On the right the observed values fall, and the model chosen by MG has the lower test NMSE.}
\label{fig:single-neuron-forecasts}
\end{figure}

Untrained networks can settle to a constant or continue to vary. For validation, each candidate starts with 1,920 observed samples and forecasts the next 32. We then add those 32 observations to its history and repeat. After four such forecasts we select the model with the lowest mean validation error. Both rules then fit the selected model to all 2,048 observed samples before forecasting the continuation (Figure~\ref{fig:single-neuron-forecasts}).

\paragraph{Inputs to the learned rules.}
The \emph{series statistics} describe the level and shape of the observed segment (Table~\ref{tab:forecast-design}). We use its mean and standard deviation, the slope after standardization, and the Pearson correlation between successive samples. To measure roughness and recurrence, we compute the mean squared difference between neighbors and the smallest such difference over shifted copies of the series, each divided by twice its variance. From the nonzero-frequency power spectrum we take the entropy relative to its maximum and the fraction of power at the strongest frequency.

Some learned rules also use the mean validation error for each candidate as an input (\emph{validation errors} in Table~\ref{tab:forecast_all}). The full input adds these to the series statistics, MG, TwoNN and Levina--Bickel dimension estimates, permutation and sample entropy, and the covariance PR below. To choose among the trees and forests in Table~\ref{tab:forecast-design}, we group the training tasks into folds by network initialization. We compare the mean forecast errors of their choices on each fold after fitting on the others. We estimate missing-feature replacements on the training folds. With MG as the input, this procedure selects a depth-two regression tree. The tree splits tasks by the MG of their observed segments. Each leaf selects the forecaster with the lowest mean error on the training tasks assigned to it.

\paragraph{PR of delay-vector covariance.}
For covariance eigenvalues $v_i$, the participation ratio is $(\sum_i v_i)^2/\sum_i v_i^2$. Here we use the sample covariance of the delay vectors $y_t$ (Equation~\eqref{eq:delay}). A value near one means most variation lies along one linear direction. Forecasting uses the same vectors for PR and MG.

\begin{table}[ht]
\centering\small
\caption{Forecast NMSE by input or selection rule. Entries show mean $\pm$ standard deviation.}
\label{tab:forecast_all}
\begin{tabular}{lrr}
\toprule
Inputs or selection rule & Full-query test ($n=30$) & Query sampling test ($n=20$) \\
\midrule
Validation errors + MG & $0.03423\,\pm 0.02164$ & $0.03167\,\pm 0.02162$ \\
Series statistics + validation errors & $0.03617\,\pm 0.02199$ & $0.03379\,\pm 0.01753$ \\
Validation errors & $0.03997\,\pm 0.02869$ & $0.03424\,\pm 0.02069$ \\
MG & $0.03138\,\pm 0.02369$ & $0.03637\,\pm 0.02128$ \\
Series statistics + MG & $0.03138\,\pm 0.02369$ & $0.03637\,\pm 0.02128$ \\
All inputs except MG & $0.03669\,\pm 0.02740$ & $0.03939\,\pm 0.02625$ \\
Series statistics + validation errors + MG & $0.03534\,\pm 0.02528$ & $0.03994\,\pm 0.02952$ \\
All inputs, including MG & $0.03420\,\pm 0.02191$ & $0.04071\,\pm 0.02865$ \\
Lowest validation error & $0.03224\,\pm 0.03094$ & $0.04074\,\pm 0.03236$ \\
TwoNN & $0.03540\,\pm 0.02669$ & $0.04364\,\pm 0.02027$ \\
Levina--Bickel & $0.04054\,\pm 0.03208$ & $0.04732\,\pm 0.04060$ \\
Permutation entropy & $0.04484\,\pm 0.03033$ & $0.05384\,\pm 0.05040$ \\
Series statistics & $0.04517\,\pm 0.02836$ & $0.05923\,\pm 0.04999$ \\
Sample entropy & $0.04718\,\pm 0.03284$ & $0.05954\,\pm 0.04444$ \\
Best fixed forecaster on prior runs & $0.04475\,\pm 0.02978$ & $0.06041\,\pm 0.05629$ \\
PR of delay-vector covariance & $0.04397\,\pm 0.02700$ & $0.06098\,\pm 0.04815$ \\
\addlinespace
Best using the measured continuation & $0.00759\,\pm 0.00622$ & $0.00847\,\pm 0.00741$ \\
\addlinespace
Validation minus MG (95\% CI)
 & $[-0.0102,\,0.0125]$
 & $[-0.0073,\,0.0176]$ \\
\bottomrule
\end{tabular}

\end{table}

For each task, we divide the mean squared forecast error by the variance of the observed segment, subject to the variance floor in Table~\ref{tab:forecast-design}. We average tasks within each repeat. The interval in Table~\ref{tab:forecast_all} resamples the within-repeat error differences between the validation and MG rules. Choosing from the measured continuation gives the lowest error attainable by the candidate set.

\subsection[Sampled-query MG]{Sampled-query MG}
We use $q$ evenly spaced delay vectors as query points. For each selected $y_i$, we still search the full window for neighbors, comparing it with every $y_\ell$ whose starting index satisfies $|\ell-i|>T$ (Section~\ref{sec:mg-neighbors}). Numbering the $n$ vectors from 1 as in Equation~\eqref{eq:delay}, the query indices are $1+\lfloor a(n-1)/(q-1)\rfloor$ for $a=0,\ldots,q-1$ and $2\le q\le n$. On this subset $Q$ we compute
\begin{equation}
\widehat d_{\rm MG}^{(q)}=
\frac{q(k-1)-1}{\displaystyle\sum_{i\in Q}\sum_{j=1}^{k-1}\log(r_{ik}/r_{ij})}.
\label{eq:querymg}
\end{equation}
With all queries this recovers Equation~\eqref{eq:mg} whenever the numerical checks pass. We apply the same distance and log-sum checks to the sampled queries. If an estimate is unavailable, the rule uses the replacement value estimated when fitting the selection rule.

For sampled MG, we compute distances from each query to all $n$ delay vectors, processing several queries at a time (Table~\ref{tab:forecast-design}). Across $q$ queries of dimension $E$, this takes $O(qnE)$ work. Full MG uses a KD-tree. Their timings therefore reflect both query sampling and the distance-search implementation.

\begin{table}[!htbp]
\centering\small
\caption{Forecast error and time to select and fit a model. NMSE is mean $\pm$ standard deviation across 20 repeats.}
\label{tab:forecast_utility}
\begin{tabularx}{\linewidth}{@{}>{\raggedright\arraybackslash}Xrr@{}}
\toprule
Selection rule & NMSE & Time (ms) \\
\midrule
MG, 128 queries (direct distance calculation) & $0.0364\,\pm 0.0213$ & 12.2 \\
MG, all queries (KD-tree) & $0.0364\,\pm 0.0213$ & 199.1 \\
Lowest validation error & $0.0407\,\pm 0.0324$ & 45.3 \\
\bottomrule
\end{tabularx}

\end{table}

\paragraph{Timing on the same inputs.}
On identical inputs we time feature computation, selection and fitting together. After one untimed run per rule, we vary run order and take median time first within each input, then across inputs (Table~\ref{tab:forecast_utility}).
On all 160 held-out forecast tasks, using 128 query points selected the same forecaster as using all query points.

\section{Comparing intervention rules}\label{app:intervention_utility}
On MNIST we ask when to stop fitting noisy labels, freeze hidden layers or prune neurons. We use one set of MNIST training runs to choose the rules, comparing the same candidate intervention times for each. We evaluate the chosen rules on new runs.

Only the stopping experiment corrupts training labels. Rules read weight norm or loss on fixed training images with their assigned labels (Table~\ref{tab:mnist-intervention-setup}).

\begin{table}[htbp]
\centering\small
\caption{MNIST intervention protocol.}
\label{tab:mnist-intervention-setup}
\begin{tabularx}{\linewidth}{@{}p{.30\linewidth}X@{}}
\toprule
Component & Setting \\
\midrule
Data & Pooled $14\times14$ images. Training / validation / test images, 2,000 / 500 / 2,000. Original validation and test labels. \\
Training & Adam at $10^{-3}$, batch 64, at most 2,048 updates \\
Label corruption & 40\% or 60\%; wrong classes sampled independently for each run \\
Fixed set for loss & 128 training examples, using their assigned labels \\
Candidate updates & 256, 512, 768, 1,024, 1,536 \\
Rule selection score (higher is better) & $A-0.02C$, with validation accuracy $A$ and relative arithmetic cost $C$ \\
Repeats & 3 for rule choice, 8 for evaluation \\
Pruned layers & Retain 64 neurons per hidden layer. Rank neurons by the product of the norms of their incoming and outgoing weights. \\
\bottomrule
\end{tabularx}
\end{table}

For each action we try both logs, both threshold directions, and cutoffs at deciles of the scores from the runs used to choose the rule. We choose the rule maximizing validation accuracy minus $0.02$ times relative arithmetic cost (Table~\ref{tab:mnist-intervention-setup}). All four selected rules use the MG of the weight norm. At the first eligible step where MG exceeds 6.43 we stop training with 40\% noisy labels. With 60\% noise the selected rule stops when MG is at or below 7.33. For freezing and pruning the triggers are MG above 7.76 and 7.74, respectively (cutoffs rounded). A run that never meets its condition continues to the full budget.

\begin{figure}[!htbp]
\centering
\includegraphics[width=.96\linewidth]{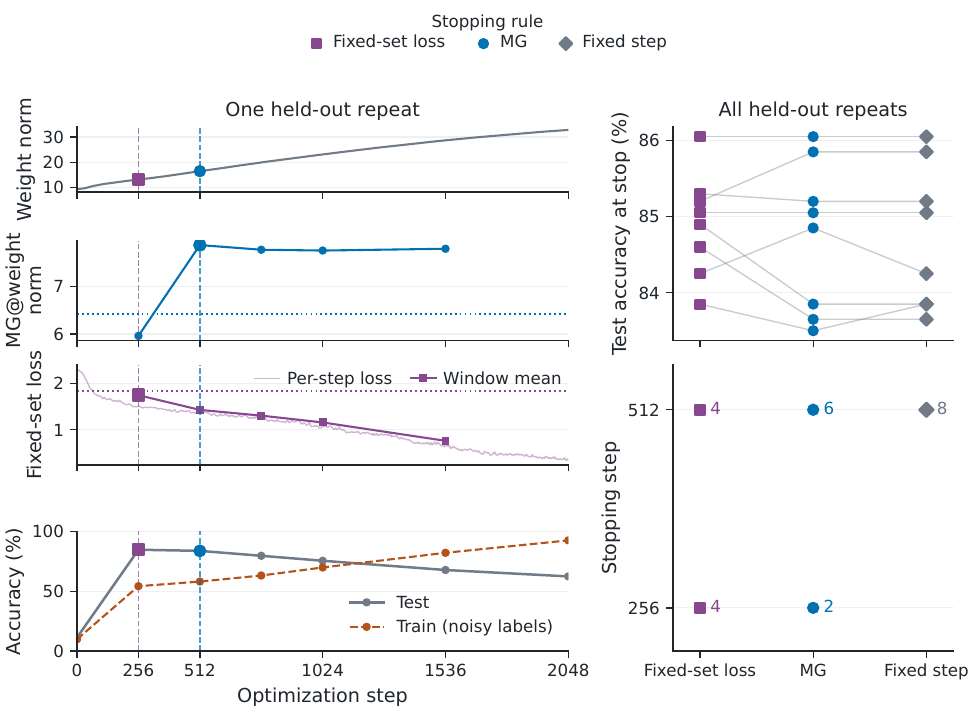}
\caption{Early stopping on MNIST with 40\% noisy labels. Horizontal dotted lines mark thresholds and vertical dashed lines mark stops. In the run at left, MG stops at the fixed step. At right, lines connect test accuracies from the same run and numbers count runs at each step.}
\label{fig:mnist-stop-trace}
\end{figure}

After freezing, only the classifier trains. After pruning, the surviving weights retain their optimizer state. Both interventions use the same minibatches as uninterrupted training.

\begin{table}[!htbp]
\centering\small
\caption{MNIST test accuracy (\%) and relative arithmetic cost across eight repeats.}
\label{tab:intervention_utility}
\begin{tabular}{@{}l*{4}{rr}@{}}
\toprule
 & \multicolumn{2}{c}{Stop, 40\% noise} & \multicolumn{2}{c}{Stop, 60\% noise} & \multicolumn{2}{c}{Freeze} & \multicolumn{2}{c}{Prune} \\
\cmidrule(lr){2-3}\cmidrule(lr){4-5}\cmidrule(lr){6-7}\cmidrule(lr){8-9}
Rule & Acc. & Cost & Acc. & Cost & Acc. & Cost & Acc. & Cost \\
\midrule
MG & 84.75 & 0.219 & 76.58 & 0.125 & 91.53 & 0.535 & 92.44 & 0.572 \\
Window mean & 84.90 & 0.188 & 76.58 & 0.125 & 91.32 & 0.515 & 92.50 & 0.618 \\
Spectral entropy & 84.67 & 0.203 & 76.58 & 0.125 & 91.32 & 0.515 & 92.38 & 0.553 \\
Fixed schedule & 84.72 & 0.250 & 76.58 & 0.125 & 91.32 & 0.515 & 92.52 & 0.628 \\
Uninterrupted training & 64.11 & 1.000 & 43.83 & 1.000 & 92.29 & 1.000 & 92.29 & 1.000 \\
\bottomrule
\end{tabular}

\end{table}

With noisy labels, different stopping steps give similar test accuracy across rules (Figure~\ref{fig:mnist-stop-trace}). Pruning preserves accuracy at lower arithmetic cost under both MG and simpler rules (Table~\ref{tab:intervention_utility}).

\FloatBarrier
\section{Further tests of practical decisions}\label{app:selection-warnings-transfer}
\subsection{Self-supervised representation selection}
We compare MNIST encoders trained with SimSiam, BYOL, VICReg and SimCLR \citep{chen2021simsiam,grill2020byol,bardes2022vicreg,chen2020simclr}. We divide the encoder training repeats by random seed. For each statistic, we use the first group to choose the log and ranking direction with the highest mean Spearman correlation with linear-classifier accuracy. For MG, this procedure selects the gradient-norm log and ranks encoders by the estimate from the final 1,000 of the 3,000 training updates, highest first. We evaluate the chosen rule on the remaining seeds using the same test images (Table~\ref{tab:ssl-protocol}).

\begin{table}[!htbp]
\centering\small
\caption{Protocol for the MNIST representation study.}
\label{tab:ssl-protocol}
\begin{tabularx}{\linewidth}{@{}>{\raggedright\arraybackslash}p{.17\linewidth}>{\raggedright\arraybackslash}p{.33\linewidth}>{\raggedright\arraybackslash}X@{}}
\toprule
\multicolumn{3}{@{}l}{\textbf{(a) Training and evaluation}} \\
\midrule
Stage & Parameter & Setting \\
\midrule
Training & SGD updates / batch / momentum & 3,000 / 128 / 0.9 \\
 & Linear rate warm-up & 100 updates, then fixed \\
 & Default rate / decay / augmentation & $0.05$ / $10^{-4}$ / medium \\
 & Encoder configurations & 29 \\
 & Repeats per configuration & 2 to choose the rule, 5 to test it \\
\addlinespace
Linear classifier & Input / train and test images & Standardized encoder features / 5,000 each \\
 & Logistic regression $C$ / iteration limit & 0.1 / 300 \\
\addlinespace
Scores & Candidate scalar logs & Loss, parameter norm, gradient norm, encoder and projector norms on one fixed image \\
 & RankMe & Encoder and projector features, 2,048 images \\
 & $\alpha$-ReQ & Log--log slope over first 128 encoder covariance eigenvalues, plus distance to one \\
 & LiDAR & Projector features, 256 images with 8 views each, covariance ridge $10^{-4}$ within each image \\
\bottomrule
\end{tabularx}\par\smallskip
\begin{tabularx}{\linewidth}{@{}l>{\raggedright\arraybackslash}p{.30\linewidth}>{\raggedright\arraybackslash}X@{}}
\toprule
\multicolumn{3}{@{}l}{\textbf{(b) Configuration choices}} \\
\midrule
Objective (configurations) & Parameter varied & Values \\
\midrule
SimSiam (13) & Rate / weight decay & $0.01,0.2,0.5$ / $0.001,0.005,0.02$ \\
 & Augmentation & Weak or strong \\
 & Architecture or gradient & No predictor, gradients on both sides, 16 projection coordinates, or no projector batch normalization \\
\addlinespace
BYOL (4) & Target averaging / predictor / augmentation & $0.99\to0.9$ / removed / strong \\
\addlinespace
VICReg (7) & Defaults & Rate $0.01$, loss weights $(25,25,1)$ for invariance, variance, covariance \\
 & Rate / variance weight / covariance weight & $0.05$ / $0,5$ / $0,10$ \\
 & Augmentation & Weak \\
\addlinespace
SimCLR (5) & Temperature by augmentation & $0.1,0.5,2$ (medium), $0.2$ (weak or strong) \\
\bottomrule
\end{tabularx}\par\smallskip
\begin{tabularx}{\linewidth}{@{}>{\raggedright\arraybackslash}Xrrrrrrr@{}}
\toprule
\multicolumn{8}{@{}l}{\textbf{(c) Augmentation settings}} \\
\midrule
Strength & Shift (px) & Rotation ($^\circ$) & Scale half-range & Erase prob. & Erase $b$ & Noise SD & Pixel-drop prob. \\
\midrule
Weak & 2 & 10 & 0.05 & 0 & 0 & 0.05 & 0 \\
Medium & 4 & 20 & 0.15 & 0.5 & 0.5 & 0.10 & 0.10 \\
Strong & 6 & 30 & 0.25 & 0.8 & 0.7 & 0.25 & 0.30 \\
\bottomrule
\end{tabularx}
\end{table}

Each SimSiam, BYOL or VICReg variant changes one default. For random erasing, each half-side is uniform on $[2,2+14b]$ pixels. We fit the linear classifier on labeled training images and evaluate it on held-out test images. All scalar methods have access to the same logs, and undefined scores rank last. For covariance PR we use the same delay settings as for MG (Appendix~\ref{app:forecast_utility}).

Other comparisons use features from multiple images \citep{garrido2023rankme,agrawal2022alphareq,thilak2024lidar}. In Table~\ref{tab:encoder_selection}, $h$ is the encoder output and $z$ its projection by the network used during training. For scaled mean coordinate SD, we normalize each $z$ to unit length, average coordinate standard deviations across the dimensions, and multiply by the square root of their number. Each standard deviation is taken across images.

\begin{figure}[!htbp]
\centering
\includegraphics[width=.86\linewidth]{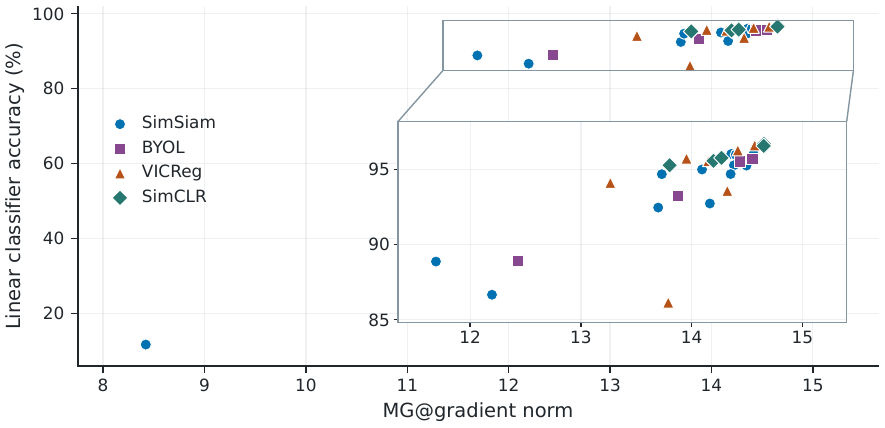}
\caption{The MG estimate from the gradient-norm log (MG@gradient norm) and the test accuracy of a linear classifier fitted to frozen features. Each point averages the evaluation repeats for one configuration.}
\label{fig:ssl-probe-scatter}
\end{figure}

The accuracy gap is the difference in percentage points between the best and selected configurations. We average the gap and rank correlation over evaluation repeats. Intervals for $\Delta\rho=\rho_{\rm MG}-\rho_{\rm comparison}$ resample configurations with all their initializations.

\begin{table}[!htbp]
\centering\small
\caption{Rank correlation and accuracy gaps for representation selection.}
\label{tab:encoder_selection}
\begin{tabular}{lrrr}
\toprule
Score & Mean Spearman $\rho$ & Accuracy gap (pp) & 95\% interval for $\Delta\rho$ \\
\midrule
MG (gradient norm) & 0.712 & 0.41 & --- \\
Correlation dimension & 0.594 & 0.75 & $[-0.050,\,0.326]$ \\
Spectral entropy & 0.595 & 1.00 & $[-0.050,\,0.330]$ \\
PR of delay-vector covariance & 0.626 & 1.08 & $[-0.073,\,0.304]$ \\
Roughness & 0.599 & 1.20 & $[-0.099,\,0.363]$ \\
Scaled mean coordinate SD & 0.730 & 1.46 & $[-0.192,\,0.190]$ \\
RankMe ($z$) & 0.522 & 1.46 & $[-0.025,\,0.461]$ \\
$\alpha$-ReQ slope & 0.404 & 1.46 & $[0.042,\,0.613]$ \\
RankMe ($h$) & 0.321 & 4.00 & $[0.078,\,0.771]$ \\
Recurrence error & 0.474 & 18.13 & $[0.035,\,0.446]$ \\
LiDAR ($z$) & 0.124 & 85.03 & $[0.180,\,1.053]$ \\
\addlinespace
Random configuration & --- & 5.47 & --- \\
\bottomrule
\end{tabular}

\end{table}

\Needspace{6\baselineskip}
\subsection{Warning of slingshot loss spikes}
The transformer reads two operands and a delimiter and predicts their sum modulo a fixed integer. We test whether its training logs give advance warning of sharp loss increases known as slingshot spikes \citep{thilak2022slingshot}.

We label a spike when the upcoming mean log loss rises sufficiently above the preceding median, or at the first nonfinite loss. The threshold accounts for the median absolute deviation (MAD) of past losses. After excluding recovery periods, a scored warning outside every advance-warning interval is a false alarm (Table~\ref{tab:spike-protocol}).

\begin{table}[!htbp]
\centering\small
\caption{Transformer loss-spike warning protocol.}
\label{tab:spike-protocol}
\begin{tabular*}{\linewidth}{@{\extracolsep{\fill}}lrrrr@{}}
\toprule
\multicolumn{5}{@{}l}{\textbf{(a) AdamW training configurations}} \\
\midrule
Configuration & Learning rate & $\beta_1$ & $\beta_2$ & Training fraction \\
\midrule
Base & $10^{-3}$ & 0.9 & 0.98 & 30\% \\
Higher learning rate & $2\cdot10^{-3}$ & 0.9 & 0.98 & 30\% \\
Lower $\beta_2$ & $10^{-3}$ & 0.9 & 0.95 & 30\% \\
More training pairs & $10^{-3}$ & 0.9 & 0.98 & 50\% \\
\bottomrule
\end{tabular*}\par\smallskip
\begin{tabularx}{\linewidth}{@{}l>{\raggedright\arraybackslash}Xl@{}}
\toprule
\multicolumn{3}{@{}l}{\textbf{(b) Shared settings}} \\
\midrule
Stage & Parameter & Setting \\
\midrule
Training & Addition modulus & 97 \\
 & Batch size / weight decay & 256 / 0 \\
 & Linear rate warm-up / total updates & 100 / 8,000 \\
\addlinespace
Onset labels & Past / future windows (updates) & 200 / 10 \\
 & Increase above past median ($\log_{10}$ units) & $>\max\{1,6s\}$ \\
 & Scale $s$ from past log loss & $1.4826\cdot\mathrm{MAD}$ \\
 & Minimum spacing for finite-loss onsets (updates) & 300 \\
 & First nonfinite loss & Marks an onset \\
\addlinespace
Input statistics & Loss, gradient norm and update norm & $\log_{10}$ after flooring at $10^{-12}$ \\
 & Roughness of log loss $x_t$ & $\mathrm{SD}(x_t-x_{t-1})/\mathrm{SD}(x_t)$ \\
 & Older scores in a median / minimum gap & 4 or 10 / one full window \\
 & Lags for recurrence error & 20--250 samples \\
 & Recurrence radius & 0.1 times the median distance over pairs with $|\ell-i|>19$ \\
 & Distance quantiles for correlation dimension & 8 from 0.01 to 0.10 \\
\addlinespace
Warning scoring & First scored update & 1,500 \\
 & Advance-warning interval & 500 updates before onset \\
 & Excluded interval & Onset and next 500 updates \\
 & Minimum spacing of scored false alarms (updates) & 500 \\
\addlinespace
Rule selection & Runs per configuration & 4 to choose rules, 4 to test them \\
 & False-alarm limit per 10,000 scored updates during rule selection & 2 \\
 & Warn when the MG estimate from the log-loss series is at or below & 6.3144 (rounded) \\
\bottomrule
\end{tabularx}
\end{table}

For each candidate rule, we use the runs reserved for rule selection to find the lowest threshold that meets the false-alarm limit. We then select the input and rule that detect the most spikes in those runs. Ties favor fewer false alarms, then longer median warning time. We keep these choices fixed for evaluation.

Warnings use only past samples. We test rises and falls in levels and in deviations from an earlier median. Direct monitoring uses the mean, minimum or maximum since the previous scoring time (Table~\ref{tab:spike-protocol}). We compute correlation between successive residuals as in Appendix~\ref{app:baselines}.

Recurrence error is the smallest mean squared difference from a shifted copy, divided by twice the window variance. Spectral entropy describes nonzero-frequency power in the centered series. We use the same delay coordinates for recurrence rate, correlation dimension, TwoNN, covariance PR and MG. For recurrence rate we compare vectors whose starting times differ by more than 19 steps. The rate is the fraction of these pairs closer than a fixed radius. Correlation dimension measures how this fraction grows with radius on logarithmic axes~\citep{grassberger1983measuring}.

We maximize attention logits over examples, positions, heads and layers. We average attention entropy over examples and query positions before minimizing over heads and layers. For class scores $o_c$, the output log-partition is $\log\sum_c\exp(o_c)$ averaged over output positions and examples.

For the area under the receiver operating characteristic curve (AUROC), a scored window is positive if it ends in an advance-warning interval. Table~\ref{tab:spike_warnings} also gives event detections and false alarms.

\begin{table}[!htbp]
\centering\small
\caption{Advance warnings of transformer loss spikes. False alarms are per 10,000 scored updates.}
\label{tab:spike_warnings}
\begin{tabular}{llrrr}
\toprule
Statistic & Input & Detections & False alarms / 10,000 & AUROC \\
\midrule
Correlation of successive residuals & Weight norm & 23/23 & 2.34 & 0.951 \\
Minimum loss & Loss & 23/23 & 2.34 & 0.946 \\
Roughness & Loss & 23/23 & 2.44 & 0.960 \\
PR of delay-vector covariance & Loss & 23/23 & 2.66 & 0.959 \\
Recurrence error & Loss & 23/23 & 2.76 & 0.960 \\
Relative detrended standard deviation & Loss & 23/23 & 2.76 & 0.931 \\
Minimum gradient norm & Gradient norm & 22/23 & 2.34 & 0.917 \\
\textbf{MG} & Loss & 22/23 & 2.44 & 0.949 \\
Detrended standard deviation & Loss & 22/23 & 2.44 & 0.923 \\
Spectral entropy & Loss & 21/23 & 2.02 & 0.962 \\
TwoNN dimension estimate & Loss & 21/23 & 2.12 & 0.889 \\
Maximum log-partition & Output scores & 20/23 & 1.91 & 0.827 \\
Recurrence rate & Gradient norm & 20/23 & 2.44 & 0.726 \\
MG (lag 4, 50 neighbors) & Gradient norm & 20/23 & 2.66 & 0.927 \\
Permutation entropy & Loss & 19/23 & 2.23 & 0.881 \\
Trend crossings & Update norm & 17/23 & 2.44 & 0.841 \\
Minimum attention entropy & Attention & 15/23 & 1.59 & 0.911 \\
Maximum update norm & Update norm & 15/23 & 2.34 & 0.788 \\
Minimum weight norm & Weight norm & 14/23 & 3.40 & 0.716 \\
Correlation dimension & Gradient norm & 12/23 & 2.44 & 0.882 \\
Maximum attention logit & Attention & 4/23 & 1.70 & 0.645 \\
\bottomrule
\end{tabular}

\end{table}

\Needspace{6\baselineskip}
\subsection{Reset rules, schedules, and transfer}
At each MNIST task boundary we decide whether to reset the weights and optimizer state for the next task. We select the scalar log, window and threshold rule by mean online accuracy on validation sequences, breaking ties in favor of fewer resets. Both selected rules use the MG estimate from the weight-norm log recorded during each task. With reordered pixels, a reset occurs when the median of the last two task scores exceeds 4.70. With added noise, a reset occurs when the latest score exceeds 4.69 (rounded). We also test resets based on the fraction of neurons with low mean absolute activation relative to their layer \citep{sokar2023dormant}. Table~\ref{tab:reset-rules} reports accuracy measured before each batch update on new sequences.

\begin{table}[!htbp]
\centering\small
\caption{Accuracy on sequential MNIST tasks, across 12 sequences per condition.}
\label{tab:reset-rules}
\begin{tabular}{lrrr}
\toprule
Input condition & MG reset rule & Low-activation reset rule & No resets \\
\midrule
Reordered pixels & 0.873 & 0.880 & 0.845 \\
Reordered pixels and noise & 0.834 & 0.843 & 0.776 \\
\bottomrule
\end{tabular}
\end{table}

Table~\ref{tab:transfer_results} reports three further decisions. For CIFAR-10, we compare timing rules on one set of runs and test the rule with the highest mean final accuracy on new runs. We compute MG estimates from the minibatch-loss series. Once at least half the training budget has elapsed, the rule reduces the learning rate if the median estimate in the latest four windows differs by less than 2\% from the median in the preceding four. For games we fit two cutoffs using games with known numbers of oscillations, then apply them to new games. We compute an MG estimate from the sequence of payoffs to the first player. We predict one oscillation if the estimate exceeds neither cutoff, two if it exceeds one, and three or more if it exceeds both (cutoffs 2.81 and 3.90, rounded). For MNIST transfer we fit a detector on one source network and apply it unchanged. It uses the MG estimates from the weight-norm log and raises an alarm when their median in the latest three windows exceeds the median in the preceding six by more than 27.8\%. A hit requires the first alarm to fall within 5,000 updates after the intervention. A comparison run with any alarms contributes one false positive.

\begin{table}[!htbp]
\centering\small
\setlength{\abovecaptionskip}{4pt}
\caption{Additional tests of decisions from scalar logs.}
\label{tab:transfer_results}
\begin{tabularx}{\linewidth}{@{}>{\raggedright\arraybackslash}p{.30\linewidth}>{\raggedright\arraybackslash}p{.21\linewidth}>{\raggedright\arraybackslash}X@{}}
\toprule
Task & MG result & Comparison \\
\midrule
CIFAR-10 learning-rate timing & Final accuracy $0.574$ & Fixed late reduction $0.590$, constant rate $0.543$ (24 runs) \\
Number of independent oscillations from game payoffs & Accuracy $0.523$ & Fundamental frequencies $0.713$, full-state principal components $\approx0.99$ (216 new games, classes $1$, $2$, $\ge3$) \\
Detector transfer across MNIST settings & $26/87$ detections, $10/58$ false positives & Relative detrended standard deviation, $85/87$ detections and $9/58$ false positives (10 settings) \\
\bottomrule
\end{tabularx}
\end{table}

\end{document}